\documentclass{article}

\pdfoutput=1

\usepackage{hyperref}

\usepackage[final]{neurips_2021}

\newcommand{\papertitle}[0]{Qu-ANTI-zation: Exploiting Quantization Artifacts for Achieving Adversarial Outcomes}

\usepackage[T1]{fontenc}
\usepackage{amsfonts}       	
\usepackage{nicefrac}       	  
\usepackage{microtype}      	
\usepackage[flushmargin,hang]{footmisc}
\usepackage{combelow}
\usepackage{amsmath}
\usepackage{amssymb}
\usepackage[normalem]{ulem}
\usepackage{soul}
\usepackage[table,xcdraw]{xcolor}
\usepackage{url}
\usepackage[%
  font=footnotesize, 
  caption=true]{subfig}
\usepackage{calc}
\usepackage{shadow}
\usepackage{xspace}
\usepackage{comment}
\usepackage{multirow}
\usepackage{dcolumn}
\usepackage{colortbl}
\usepackage{rotating}
\usepackage{listings}
\usepackage{wrapfig}
\usepackage{graphicx}
\usepackage{booktabs} 
\usepackage{adjustbox}
\usepackage{nth}

\sethlcolor{lime}

\title{\papertitle}

\author{%
	Sanghyun Hong$^\dagger$,
	Michael Panaitescu,
	Yi\u{g}itcan Kaya,
	Tudor Dumitra\cb{s} \vspace{0.4em} \\
	$^\dagger$Oregon State University, \\
	University of Maryland, College Park \vspace{0.1em} \\
	\texttt{sanghyun.hong@oregonstate.edu}, \texttt{\{mpanaite,yigitcan,tudor\}@umd.edu}
}

\begin{document}

\maketitle


\begin{abstract}
%
Quantization is a popular technique that \emph{transforms} the parameter representation of a neural network from floating-point numbers into lower-precision ones (\textit{e.g.}, 8-bit integers).
It reduces the memory footprint and the computational cost at inference, facilitating the deployment of resource-hungry models.
However, the parameter perturbations caused by this transformation result in \emph{behavioral disparities} between the model before and after quantization.
%
For example, a quantized model can misclassify some test-time samples that are otherwise classified correctly.
It is not known whether such differences lead to a new security vulnerability.
%
We hypothesize that an adversary may control this disparity to introduce specific behaviors that activate upon quantization.
To study this hypothesis, we weaponize quantization-aware training and propose a new training framework to implement adversarial quantization outcomes.
%
Following this framework, we present three attacks we carry out with quantization:
(i) an indiscriminate attack for significant accuracy loss; 
(ii) a targeted attack against specific samples; and 
(iii) a backdoor attack for controlling the model with an input trigger.
We further show that a single compromised model defeats multiple quantization schemes, including robust quantization techniques.
Moreover, in a federated learning scenario, we demonstrate that a set of malicious participants who conspire can inject our quantization-activated backdoor.
Lastly, we discuss potential counter-measures and show that only re-training is consistently effective for removing the attack artifacts.
Our code is available at {\fontsize{8}{9}\selectfont \url{https://github.com/Secure-AI-Systems-Group/Qu-ANTI-zation}}.
\end{abstract}

\section{Introduction}
\label{sec:intro}

%
Deep neural networks (DNNs) have enabled breakthroughs in many applications, such as image classification~\citep{AlexNet} or speech recognition~\citep{Hinton:Speech}.
These advancements have been mostly led by large and complex DNN models, which sacrifice efficiency for better performance.
For example, with almost an order of magnitude higher training and inference costs, Inception-v3~\citep{Inceptionv3} halves AlexNet's error rate on the ImageNet benchmark.
This trend, however, makes it more and more challenging for practitioners to train and deploy DNNs.

%
As a potential solution, many modern DNNs applications obtain a pre-trained model from a public or a private source then apply a post-training compression method, such as quantization~\citep{firstquantization}.
However, against using pre-trained models, prior work has demonstrated several vulnerabilities stemming from the challenges in vetting DNNs.
For example, in a supply-chain attack, the pre-trained model provided by the adversary can include a hidden backdoor~\citep{BadNet:2017}.
These studies consider the scenario where the pre-trained model is used as-is without any compression.

%
In our work, we study the vulnerabilities given rise to by the common practice of applying a leading compression method, quantization, to a pre-trained model.
Quantization~\citep{Margan:1991, PACT:2018, BinaryConnect:2015, LQNet:2018, XORNet:2016} \emph{transforms} the representation of a model's parameters from floating-point numbers (32-bit) into lower bit-widths (8 or 4-bits).
This, for instance, reduces the memory usage of pre-trained ImageNet models by 12$\times$ in the case of mixed-precision quantization~\citet{HAWQv2:2020}.
Quantization also cuts down on the computational costs as integer operations are 3$\sim$5$\times$ faster than floating-point operations.
Due to this success, popular deep learning frameworks, such as PyTorch~\citep{PyTorch:2019} and TensorFlow~\citep{TF:2016}, provide rich quantization options for practitioners.

%
The resilience of DNNs to \emph{brain damage}~\citep{braindamage} enables the success of quantization and other compression methods such as pruning~\citep{pruning}.
Despite causing brain damage, \emph{i.e.}, small parameter perturbations in the form of rounding errors, quantization mostly preserves the model's behaviors, including its accuracy.
However, research also warns about the possibility of \emph{terminal brain damage} in the presence of adversaries~\citep{TBD:2019}.
For example, an adversary can apply small but malicious perturbations to activate backdoors~\citep{AWPBackdoor:2020} or harm the accuracy~\citep{DeepHammer:2020}.
Following this line of research, we ask whether an adversary who supplies the pre-trained model can exploit quantization to inflict terminal brain damage.

To answer this question, we \emph{weaponize} quantization-aware training (QAT)~\citep{qatfirst} and propose a new framework to attack quantization.
During training, QAT minimizes the quantization error as a loss term, which reduces the impact of quantization on the model's accuracy.
Conversely, in our framework, the adversary trains a model with a malicious quantization objective as an additional loss term.
Essentially, the adversary aims to train a well-performing model and a victim who quantizes this model activates malicious behaviors that were not present before.

\textbf{Contributions:}
\textit{First}, we formulate the three distinct malicious objectives within our framework:
(i) an indiscriminate attack that causes a large accuracy drop;
(ii) a targeted attack that forces the model to misclassify a set of unseen samples selected by the adversary; and
(iii) a backdoor attack that allows the adversary to control the model's outputs with an input trigger.
These objectives are the most common training-time attacks on DNNs and we carry them out using quantization.

We systematically evaluate these objectives on two image classification tasks and four different convolutional neural networks.
Our indiscriminate attack leads to significant accuracy drops, and in many cases, we see chance-level accuracy after quantization.
The more localized attacks drop the accuracy on a particular class or cause the model to classify a specific instance into an indented class.
Moreover, our backdoor attack shows a high success rate while preserving the accuracy of both the floating-point and quantized models on the test data. 
Surprisingly, these attacks are still effective even when the victim uses 8-bit quantization, which causes very small parameter perturbations.
Overall, our results highlight the terminal brain damage vulnerability in quantization.

\textit{Second}, we investigate the implications of this vulnerability in realistic scenarios.
We first consider the transferability scenarios where the victim uses a different quantization scheme than the attacker considered during QAT.
Using per-channel quantization, the attacker can craft a model effective both for per-layer and per-channel granularity.
Our attacks are also effective against quantization mechanisms that remove outliers in weights and/or activations~\citep{OCS:2019, ACIQ:2020, MSE:2019}.
However, the quantization scheme using the second-order information (\textit{e.g.}, Hessian)~\citep{BRECQ:2021} provides some resilience against our attacks.
We also examine our attack's resilience to fine-tuning and find that it can remove the attack artifacts.
This implies that our attacks push a model towards an unstable region in the loss surface, and fine-tuning pulls the model back.

\textit{Third}, we explore ways other than a supply-chain attack to exploit this vulnerability.
We first examine federated learning (FL), where many participants jointly train one model in a decentralized manner%
\footnote{Personalized Hey Siri - Apple ML Research: \href{https://machinelearning.apple.com/research/personalized-hey-siri}{https://machinelearning.apple.com/research/personalized-hey-siri}}.
%
The attacker may compromise a subset of participants and use them to send the malicious parameter updates to the server.
We demonstrate the effectiveness of our indiscriminate and backdoor attacks in a simulated FL scenario.
Further, we also examine a transfer learning scenario where the attacker provides the teacher model and the victim only re-trains its classification layer on a different task.
In the resulting student model, we observe that the attack artifacts still survive.
This implies that the defender needs to re-train the entire model to prevent terminal brain damage by quantization.
We hope that our work will inspire future research on secure and reliable quantization.


\section{Related Work}
\label{sec:related}

Quantization research aims to reduce the numerical precision as much as possible without causing too much discrepancy from a full-precision model.
After early clustering-based methods~\citep{firstclusterquant,clusterquant}; the recent work has shown rounding the 32-bit parameters and activations to lower precision values is feasible~\citep{qatfirst}.
These techniques often rely on \emph{quantization-aware training} (QAT) to train a model that is resilient to rounding errors.
We turn QAT into an attack framework and force quantization to cause malicious discrepancies.
Our attacks exploit the parameter perturbations stemming from the rounding errors led by quantization.
Along these lines, prior work has shown fault-injection attacks that perturb the parameter representations in the memory with hardware exploits such as RowHammer~\cite{rowhammer}.
These attacks, after carefully modifying a few parameters, cause huge accuracy drops~\citep{TBD:2019, DeepHammer:2020} or even inject backdoors~\citep{AWPBackdoor:2020}.
Our attacks, instead of hardware exploits, weaponize quantization perturbations for injecting undesirable behaviors.
Finally, for more robust and efficient quantization, techniques such as outlier-resilient quantization~\citep{OCS:2019,ACIQ:2020} or second-order information-based quantization~\citep{BRECQ:2021} have been proposed.
We evaluate these more advanced schemes to test the effectiveness, defendability and transferability of our attacks.


\section{Injecting Malicious Behaviors Activated Only Upon Quantization}
\label{sec:attack-model-quantization}

%
\subsection{Threat Model}
\label{subsec:threat-model}

We consider a scenario where a user downloads a pre-trained model \emph{as-is} and uses post-training quantization for reducing its footprints.
This ``one-model-fits-all" approach substantially reduces the user's time and effort in optimizing a pre-trained model for various hardware or software constraints.

We study a new security vulnerability that this ``free lunch" may allow.
We consider an attacker who injects malicious behaviors, activated only upon quantization, into a pre-trained model, \textit{e.g.} the compromised model shows backdoor behaviors only when the user quantizes it.
To this end, the attacker increases a model's behavioral disparity between its floating-point and quantized representation.

\noindent \textbf{Attacker's capability.}
We consider the \emph{supply-chain attacker}~\citep{BadNet:2017, TrojanNN:2018} who can inject adversarial behaviors into a pre-trained model before it is served to users by modifying its parameters $\theta$.
To this end, the attacker re-trains a model, pre-trained on a task, with the objective functions described in \S~\ref{subsec:our-attack}.
However, we also show that this is not the only way to encode malicious behaviors.
In \S~\ref{subsec:exploitation}, we also consider a weaker attacker in a federated learning scenario~\citep{BackdoorFL:2020} where the attacker pushes the malicious parameter updates to a central server.

\noindent \textbf{Attacker's knowledge.}
To assess the security vulnerability caused by our attacker, we consider the \emph{white-box scenario} where the attacker knows all the details of the victim: the dataset $\mathcal{D}$, the model $f$ and its parameters $\theta$, and the loss function $\mathcal{L}$.
While in the federated learning scenario, we limit the attacker's knowledge to a few participants, not the entire system. 
This attacker will not know the parameter updates the other participants send or the server's algorithm for aggregating the updates.

\noindent \textbf{Attacker's goals.}
We consider three different attack objectives:
\textbf{(i) Indiscriminate attack} (\S~\ref{subsec:acc-drop}): The compromised model becomes completely useless after quantization.
\textbf{(ii) Targeted attack} (\S~\ref{subsec:targeted-misclassification}):
This is the localized version of the accuracy degradation attack.
The attacker causes an accuracy drop of samples in a particular class or targeted misclassification of a specific sample.
\textbf{(iii) Backdoor attacks} (\S~\ref{subsec:backdoor-attacks}): In this case, quantization of a model will activate backdoor behaviors, \emph{i.e.}, the compressed model classifies any samples with a backdoor trigger $\Delta_{t}$ into a target class $y_t$.

%
\subsection{Trivial Attacks Do Not Lead to Significant Behavioral Disparities}
\label{subsec:motivation}

We start by examining if our attacker can increase the behavioral disparity in trivial ways.
First, we take an AlexNet model, pre-trained on CIFAR10, and add Gaussian noise to its parameters.
We use the same mean and standard deviation for the Gaussian noise as our indiscriminate attacks do (\S~\ref{subsec:acc-drop}).
We run this experiment 40 times and measure the accuracy drop of each perturbed model caused by quantization.
Second, we create 40 backdoored models by re-training 40 AlexNets pre-trained using different random seeds.
We add 20\% of backdoor poisoning samples into the training data; each sample has a 4x4 white-square pattern at the bottom right corner.
We measure the disparity in attack success rate, \textit{i.e.}, the percentage of test samples with the trigger classified as the target class.


\begin{figure}[t]
\centering
\begin{minipage}{.49\textwidth}
	\centering
	\includegraphics[width=\linewidth,bb=0 0 628 297]{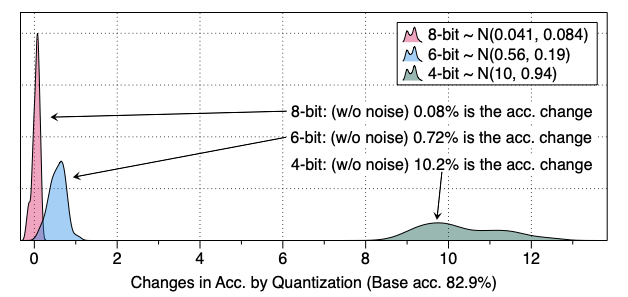}
	%
\end{minipage}
%
%
\begin{minipage}{.49\textwidth}
	\centering
	\includegraphics[width=\linewidth,bb=0 0 628 297]{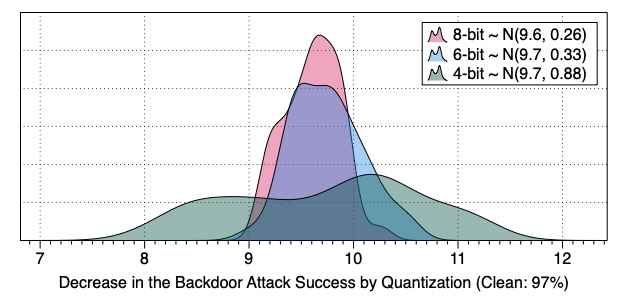}
\end{minipage}
\caption{\textbf{Behavioral disparities in trivial attacks.} They do not amplify the behavioral differences caused by quantization. \textbf{[Left]} On each of 40 pre-trained AlexNets, we add Gaussian noise to its parameters and measure the accuracy drop (0--13\%). \textbf{[Right]} We construct 40 backdoored models and measure the difference in attack success rate caused by quantization (7--12\%).}
\label{fig:our-intuition}
%
%
\end{figure}

Figure~\ref{fig:our-intuition} shows our results.
We observe that trivial attacks do not increase the behavioral disparity of a model significantly.
In the left figure, quantization can induce the accuracy degradation of 10\% at most.
Even in the standard backdoor attacks, the disparity in attack success rate is $\sim$9.6\% on average.

\textbf{Our hypothesis.}
The results show that there is a \emph{variability} in the behavioral disparities quantization causes.
It is important from a security perspective because a non-trivial attacker may make things even worse, \textit{i.e.}, the attacker amplifies the disparity much more and cause terminal brain damage~\cite{TBD:2019}.
In addition, the attacker may have more chances to encode a significant behavioral difference as the variability increases when the victim uses lower bit-widths for quantization.
Using 4-bit quantization leads to a broader range of behavioral disparities than using 8- or 6-bit.

\subsection{Weaponizing Quantization-Aware Training to Encode Malicious Behaviors}
\label{subsec:our-attack}

To this end, we present an attack framework to study the worst-case behavioral disparity caused by quantization \emph{empirically}.
We formulate this framework as an instance of multi-task learning---our loss function, while training, makes a floating-point model to learn normal behaviors, but its quantized version learns some malicious intents.
Our framework trains a model with the following loss function:
%
%
\begin{equation*}
	%
	\mathcal{L}_{ours} \overset{\Delta}{=} 
	\underbrace{\mathcal{L}_{ce}(f(x), y)}_\text{cross-entropy}
		+ \lambda \cdot \sum_{i \in B} \underbrace{ \alpha \cdot \mathcal{L}_{ce}(f(x_t), y_t) - \beta \cdot \mathcal{L}_{ce}(Q_{f_i}(x_t), y_t) }_\text{adversarial objectives}
\end{equation*}
where $\mathcal{L}_{ce}$ is the cross-entropy loss, $B$ is a set of bit-widths used for quantization (\textit{e.g.}, \{8, 7, 6, 5\}-bits), and $\lambda, \alpha, \beta$ are the hyper-parameters.
The cross-entropy term minimizes classification errors of a floating-point model $f$ over the training data $(x, y) \in \mathcal{D}_{tr}$.
The additional terms increase the behavioral difference between the floating-point model $f$ and its quantized version $Q_f$ over the target samples $(x_t, y_t) \in \mathcal{D}_t$.
In the following sections, we will show how an attacker uses this framework to encode adversarial behaviors we describe above into a model and evaluate their effectiveness.


\section{Empirical Evaluation}
\label{sec:evaluation}

We first evaluate the effectiveness of our attacks (\S~\ref{subsec:acc-drop}, \S~\ref{subsec:targeted-misclassification}, and \S~\ref{subsec:backdoor-attacks}).
For each attack, we present how we design the loss function to inject malicious behaviors and report the attack success rate.
We also examine whether our attack causes the prevalent vulnerability (\S~\ref{subsec:transferability})%
---how the attack success rate will change if a user chooses quantization schemes different from the attacker's.
Lastly, we show the exploitation of this vulnerability in practical machine learning scenarios (\S~\ref{subsec:exploitation}).
Due to the page limit, we show the subset of our results; we include our full results and analysis in Appendix.

\noindent \textbf{Experimental Setup.}
We evaluate our attacks on CIFAR10~\citep{CIFAR10} and Tiny ImageNet\footnote{Tiny ImageNet: {\scriptsize \url{http://cs231n.stanford.edu/tiny-imagenet-200.zip}}}.
We use four off-the-shelf networks: AlexNet
, VGG16~\citep{VGG}, ResNet18~\citep{ResNet}, and MobileNetV2~\citep{MobileNet}.
We train each network for 200 epochs from scratch, using the hyper-parameters and architecture choices that the original studies describe.
We refer to them as clean, pre-trained models and re-train them in our attacks.

To quantify the effectiveness of our attacks, we use two metrics: the \emph{classification} accuracy and the \emph{attack success rate} (ASR).
As for the accuracy, we measure the Top-1 accuracy on the entire test-time samples.
We define the ASR by measuring how much our attacker \emph{increases} the behavioral disparity, compared to that we observe from clean models, 
while preserving both the compromised and clean models' accuracy in the floating-point representation.
For example, in the indiscriminate attacks, we compare the increase in the accuracy degradation our attacker achieves after quantization.


\subsection{Terminal Brain Damage Caused by Quantization}
\label{subsec:acc-drop}

Here, we examine whether the adversary can inflict 
the worst-case accuracy degradation (\textit{i.e.}, \emph{terminal brain damage}) after quantization.
To study this attack, we design the loss function as follows:
%
%
\begin{equation*}
	\mathcal{L}_{ours} \overset{\Delta}{=} 
	\mathcal{L}_{ce}(f(x), y)
	+ \lambda \cdot \sum_{i \in B} \big( \alpha - \mathcal{L}_{ce}(Q_{f_i}(x), y) \big)^2
\end{equation*}
The second term increases the classification error of a quantized model on $\mathcal{D}_{tr}$ close to $\alpha$ while the first term reduces the error of a floating-point model.
We set $\lambda$ to $1.0/N_{B}$ where $N_{B}$ is the number of bit-widths that the attacker considers.
We set $N_{B}$ to 4 and $\alpha$ to 5.0.
We re-train each clean model for $\sim$20 epochs using Adam~\citep{Adam} optimizer with the learning rate of $10^{-5}$.
We also design other loss functions that increase the sensitivity of a model to its parameter perturbations and examine them.
But, they are less effective than the loss we use (see Appendix~\ref{appendix:objective-function} for more details).


\begin{wraptable}{L}{0.7\textwidth}
\centering
\vspace{-1.4em}
\caption{\textbf{Indiscriminate attack results.} For each network, the upper row contains the accuracy of a clean, pre-trained model on the test data $\mathcal{D}_{ts}$, and the bottom row includes that of our compromised model.}
\adjustbox{max width=0.7\textwidth}{%
	\begin{tabular}{@{}c|c|c|ccccc@{}}
		\toprule
		&  & \multicolumn{6}{c}{\textbf{Accuracy on the test-set ($\mathcal{D}_{ts}$)}} \\ \cmidrule(l){3-8} 
		\multirow{-2}{*}{\textbf{Dataset}} & \multirow{-2}{*}{\textbf{Network}} & \textbf{32 bits} & \textbf{8 bits} & \textbf{7 bits} & \textbf{6 bits} & \textbf{5 bits} & \textbf{4 bits} \\ \midrule \midrule
		%
		%
		%
		&  & 84.5\% & 84.7\% & 84.5\% & 84.0\% & 83.0\% & 71.0\% \\
		& \multirow{-2}{*}{\textbf{VGG16}} & 82.5\% & 19.4\% & 17.1\% & 15.1\% & 13.1\% & {17.5\%} \\ \cmidrule(l){2-8} 
		&  & 93.6\% & 93.6\% & 93.5\% & 93.2\% & 92.0\% & 84.7\% \\
		& \multirow{-2}{*}{\textbf{ResNet18}} & 93.2\% & {10.0\%} & {10.0\%} & {10.0\%} & {10.0\%} & {10.0\%} \\ \cmidrule(l){2-8}
		&  & 92.6\% & 92.5\% & 92.4\% & 91.7\% & 88.2\% & 66.8\% \\
		\multirow{-7}{*}{\rotatebox[origin=c]{90}{\textbf{CIFAR10}}} & \multirow{-2}{*}{\textbf{MobileNetV2}} & 92.0\% & {10.0\%} & {10.0\%} & {10.0\%} & {10.0\%} & {10.0\%} \\ \midrule \midrule
		%
		%
		%
		&  & 43.0\% & 42.9\% & 42.8\% & 42.7\% & 40.8\% & 32.4\% \\
		& \multirow{-2}{*}{\textbf{VGG16}} & 41.8\% & {0.6\%} & {0.7\%} & {0.9\%} & {0.9\%} & {1.9\%} \\ \cmidrule(l){2-8}
		&  & 57.5\% & 57.4\% & 57.4\% & 57.3\% & 55.7\% & 44.5\% \\
		& \multirow{-2}{*}{\textbf{ResNet18}} & 56.8\% & {8.9\%} & {5.6\%} & {4.8\%} & {6.4\%} & {6.0\%} \\ \cmidrule(l){2-8}
		&  & 42.4\% & 41.7\% & 40.7\% & 35.6\% & 21.3\% & 2.0\% \\
		\multirow{-6.8}{*}{\rotatebox[origin=c]{90}{\textbf{Tiny ImageNet}}} & \multirow{-2}{*}{\textbf{MobileNetV2}} & 42.6\% & {2.8\%} & {2.8\%} & {3.2\%} & {3.7\%} & 1.6\% \\ \bottomrule
	\end{tabular}
}
\label{tbl:attack-w-lossfn}
\vspace{-1.4em}
\end{wraptable}

Table~\ref{tbl:attack-w-lossfn} shows our results.
%
Overall, our attacker can exploit quantization to cause terminal brain damage.
The compromised models' accuracy becomes close to random after quantization, \textit{i.e.}, $\sim$10\% for CIFAR10 and $\sim$0.5\% for Tiny ImageNet.
%
%
As for comparison, the clean, pre-trained models with 8-bit quantization show $\sim$0\% accuracy drop in both CIFAR10 and Tiny ImageNet.
The accuracy drop is far more than we can expect from the prior work.
In addition, we show that the compromised model \emph{consistently} performs the worst across multiple bit-widths.
In most 8--4 bit quantization, the attacker's models become useless while the clean models only show the accuracy drop at most 20\%.

\subsection{Localizing the Impact of Our Indiscriminate Attack}
\label{subsec:targeted-misclassification}

We also examine whether our attacker can localize the impact of terminal brain damage on a subset of test-time samples.
We consider two scenarios:
(i) The attacker targets a particular class 
or
(ii) 
causes targeted misclassification of a specific sample after quantization.
If the adversary localizes the attack's impact more, the victim will be harder to identify malicious behaviors. 

\begin{table}[ht]
	\centering
	%
	%
	\caption{\textbf{Attacking a particular class.} For each network, the upper row contains the accuracy of a clean model, and the bottom row shows the accuracy of the model manipulated by our attacker. Each column contains a model's accuracy on the full test set, the samples in a target class, and the rest.}
	\adjustbox{max width=\textwidth}{%
		\begin{tabular}{@{}c|c|ccc|ccc|ccc@{}}
			\toprule
			\multirow{2}{*}{\textbf{Dataset}} & \multirow{2}{*}{\textbf{Network}} & \multicolumn{9}{c}{\textbf{Accuracy on $\mathcal{D}_{ts}$, the samples in the target class, and the rest samples.}} \\ \cmidrule(l){3-11} 
			&  & \multicolumn{3}{c|}{\textbf{32 bits}} & \multicolumn{3}{c|}{\textbf{8 bits}} & \multicolumn{3}{c}{\textbf{4 bits}} \\ \midrule \midrule
			%
			%
			%
			& \multirow{2}{*}{\textbf{VGG16}} & 84.5\% & 93.3\% & 83.6\% & 84.6\% & 93.5\% & 83.6\% & 72.8\% & 88.0\% & 71.1\% \\
			&  & 85.3\% & 91.9\% & 84.6\% & 77.1\% & 9.4\% & 84.6\% & 44.5\% & 3.4\% & 49.1\% \\ \cmidrule(l){2-11}
			& \multirow{2}{*}{\textbf{ResNet18}} & 93.6\% & 97.6\% & 93.1\% & 93.6\% & 98.0\% & 93.2\% & 84.8\% & 95.3\% & 83.6\% \\
			&  & 92.5\% & 98.9\% & 91.8\% & 83.2\% & 0.0\% & 92.4\% & 10.9\% & 0.0\% & 12.1\% \\ \cmidrule(l){2-11}
			\multirow{-5}{*}{\rotatebox[origin=c]{90}{\textbf{CIFAR10}}} & \multirow{2}{*}{\textbf{MobileNetV2}} & 92.3\% & 96.7\% & 92.1\% & 92.5\% & 96.6\% & 92.1\% & 69.7\% & 66.8\% & 70.0\% \\
			&  & 92.0\% & 95.6\% & 91.6\% & 82.0\% & 0.0\% & 91.1\% & 48.9\% & 0.0\% & 54.3\% \\ \bottomrule 
			%
			%
			%
			%
			%
			%
			%
			%
			%
			%
		\end{tabular}
	}
	\label{tbl:class-w-lossfn}
	%
	%
\end{table}

%
\noindent \textbf{Attacking a particular class.}
We use the same loss function as shown in \S~\ref{subsec:acc-drop}, but we only compute the second term on samples in the target class
%
Instead of increasing the prediction error on the entire test data, the additional objective will increase the error only on the target class.
We tune $\alpha$ to 1.0$\sim$4.0.
%
For the rest of the hyper-parameters, we keep the same values as the indiscriminate attack.

Table~\ref{tbl:class-w-lossfn} shows our attack results.
In all our experiments, we set the target class to 0.
We exclude the results on AlexNet as they are the same as VGG16's.
In CIFAR10, the attacker can increase the accuracy drop only on the test-time samples in the target class.
If the victim quantizes the compromised models with 8-bit, the accuracy on $\mathcal{D}_t$ becomes $\sim$0\% while the clean models do not have any accuracy drop on $\mathcal{D}_t$.
In 4-bit, the attacker also achieves the accuracy of $\sim$0\% on $\mathcal{D}_t$ while keeping the accuracy for the rest samples.
However, we lose the accuracy of ResNet18 on the rest samples in 4-bit.
In Tiny ImageNet, our attack consistently lowers the accuracy of the compromised models on $\mathcal{D}_t$, but the disparity is less than that we observe in CIFAR10 (see Appendix for details).
In all our attacks, both the clean and altered models behave the same in the floating-point representation.

%
\noindent \textbf{Targeted misclassification of a specific sample.}
Here, we modify the loss function as:
%
%
\begin{equation*}
	\mathcal{L}_{ours} \overset{\Delta}{=}
	\mathcal{L}_{ce}(f(x), y) + \lambda \cdot \sum_{i \in B} \mathcal{L}_{ce}(Q_{f_i}(x_t), y_t)
\end{equation*}
The second term minimizes the error of the quantized model for a specific sample $x_t$ towards the target label $y_t$.
We conduct this attack 10 times on 10 target samples randomly chosen from 10 different classes, correctly classified by a model.
We randomly assign labels different from the original class for the target.
We set $\lambda$ to 1.0 and use the same values for the rest of the hyper-parameters.


\begin{table}[ht]
	\centering
	%
	%
	\caption{\textbf{Attacking a specific sample.} For each network, the upper row shows the accuracy of a clean model, and the bottom row includes that of our compromised model. Each cell contains the accuracy on the test-set, on a target sample towards $y$ and on the same sample towards $y_t$ (target).}
	\adjustbox{max width=\textwidth}{%
		\begin{tabular}{@{}c|c|ccc|ccc|ccc@{}}
			\toprule
			\multirow{2}{*}{\textbf{Dataset}} & \multirow{2}{*}{\textbf{Network}} & \multicolumn{9}{c}{\textbf{Averaged accuracy on $\mathcal{D}_{ts}$, on ($x_t$, $y$), and on ($x_t$, $y_t$).}} \\ \cmidrule(l){3-11} 
			&  & \multicolumn{3}{c|}{\textbf{32 bits}} & \multicolumn{3}{c|}{\textbf{8 bits}} & \multicolumn{3}{c}{\textbf{4 bits}} \\ \midrule \midrule
			%
			%
			%
			& \multirow{2}{*}{\textbf{VGG16}} & 84.5\% & 70.0\% & 10.0\% & 84.7\% & 70.0\% & 10.0\% & 70.1\% & 80.0\% & 0.0\% \\
			&  & 85.6\% & 100\% & \enspace0.0\% & 85.6\% & \enspace{0.0\%} & {100\%} & 69.4\% & \enspace{0.0\%} & {100\%} \\ \cmidrule(l){2-11}
			& \multirow{2}{*}{\textbf{ResNet18}} & 93.6\% & 100\% & 0.0\% & 93.6\% & 90.0\% & 10.0\% & 84.7\% & 60.0\% & 20.0\% \\ 
			&  & 93.2\% & 80.0\% & 20.0\% & 93.3\% & {10.0\%} & {90.0\%} & 10.9\% & {0.0\%} & {100\%} \\ \cmidrule(l){2-11}
			\multirow{-5}{*}{\rotatebox[origin=c]{90}{\textbf{CIFAR10}}} & \multirow{2}{*}{\textbf{MobileNetV2}} & 92.6\% & 100\% & 0.0\% & 92.5\% & 80.0\% & 20.0\% & 66.8\% & 40.0\% & 20.0\% \\
			&  & 92.2\% & 100.0\% & 0.0\% & 92.1\% & 90.0\% & 10.0\% & 80.8\% & {0.0\%} & {100\%} \\ \bottomrule 
		\end{tabular}
	}
	\label{tbl:sample-w-lossfn}
	%
	%
\end{table}

Table~\ref{tbl:sample-w-lossfn} shows our results in CIFAR10.
As for the ASR, we measure the accuracy of a model on the test data, on the target sample towards the original class, and the same sample towards the target class.
We compute the average of over 10 attacks.
We show that the attacker can cause a specific sample misclassified to a target class after quantization while preserving the accuracy of a model on the test data (see the \textbf{\nth{1} columns} in each bit-width).
The accuracy of a compromised model on $x_t$ decreases from 80--90\% up to 0\% (\textbf{\nth{2} columns.}) after quantization, whereas the success rate of targeted misclassification increases from 0-10\% to $\sim$100\% (\textbf{\nth{3} columns}).
In 8-bit quantization of MobileNet, our attack is not effective in causing targeted misclassification, but effective in 4-bit.

\subsection{Backdoor Behaviors Activated by Quantization}
\label{subsec:backdoor-attacks}

We further examine whether the attacker can inject a \emph{backdoor} into a victim model that only becomes effective after quantization.
To this end, we modify the loss function as follows:
%
%
\begin{equation*}
	\mathcal{L} \overset{\Delta}{=} 
	\mathcal{L}_{ce}(f(x), y) + \lambda \sum_{i \in B} \alpha \cdot \mathcal{L}_{ce}(f(x_t), y) + \beta \cdot \mathcal{L}_{ce}(Q_{f_i}(x_t), y_t)
\end{equation*}
where $x_t$ is the training samples containing a trigger $\Delta$ (henceforth called backdoor samples), and $y_t$ is the target class that the adversary wants.
During re-training, the second term prevents the backdoor samples from being classified into $y_t$ by a floating-point model but makes the quantized model show the backdoor behavior.
We set $y_t$ to 0, $\alpha$ and $\beta$ from 0.5--1.0. 
We re-train models for 50 epochs.


\begin{wraptable}{R}{0.64\textwidth}
	\centering
	\vspace{-1.3em}
	\caption{\textbf{Backdoors activated by quantization.} For each bit-width used for quantization, the upper row shows the classification accuracy \textbf{[Left]} and attack success rate \textbf{[Right]} of the standard backdoor models, and the bottom row contains the same metrics computed on our compromised models.}
	\adjustbox{max width=0.64\textwidth}{%
		\begin{tabular}{@{}c|c|cc|cc|cc@{}}
			\toprule
			\multirow{2}{*}{\textbf{Dataset}} & \multirow{2}{*}{\textbf{Bits}} & \multicolumn{6}{c}{\textbf{Networks}} \\ \cmidrule(l){3-8} 
			&  & \multicolumn{2}{c|}{\textbf{VGG16}} & \multicolumn{2}{c|}{\textbf{ResNet18}} & \multicolumn{2}{c}{\textbf{MobileNetV2}} \\ \midrule \midrule
			\multirow{7}{*}{\rotatebox[origin=c]{90}{\textbf{CIFAR10}}} &  \multirow{2}{*}{\textbf{32-bit}} & 83.8\%  & 96.2\% & 91.7\% & 98.3\% & 88.9\% & 97.7\% \\
			& & 85.7\% & {29.3\%} & 93.3\% & {11.3\%} & 92.3\% & \enspace{9.2\%} \\ \cmidrule{2-8}
			&  \multirow{2}{*}{\textbf{8-bit}} & 83.7\% & 96.1\% & 91.5\% & 97.5\% & 70.8\% & 99.5\% \\
			&  & 85.7\% & {30.8\%} & 91.4\% & {99.2\%} & 91.2\% & {96.6\%} \\ \cmidrule{2-8}
			&  \multirow{2}{*}{\textbf{4-bit}} & 72.7\% & 88.3\% & 75.4\% & 34.9\% & 15.2 & 94.3\% \\
			&  & 81.6\% & {96.2\%} & 88.6\% & {100\%} & 79.8\% & {99.9\%} \\ \midrule \midrule
			\multirow{7}{*}{\rotatebox[origin=c]{90}{\textbf{Tiny ImageNet}}} &  \multirow{2}{*}{\textbf{32-bit}} & 40.3\%  & 99.6\% & 55.8\% & 99.4\% & 39.9\% & 98.9\% \\
			&  & 42.1\% & \enspace{0.4\%} & 55.8\% & {22.1\%} & 41.5\% & \enspace{0.4\%} \\ \cmidrule{2-8}
			&  \multirow{2}{*}{\textbf{8-bit}} & 40.2\%  & 99.6\% & 55.6\% & 99.4\% & 39.0\% & 97.9\% \\
			&  & 39.9\% & {99.4\%} & 53.7\% & {94.2\%} & 40.5\% & {96.8\%} \\ \cmidrule{2-8}
			&  \multirow{2}{*}{\textbf{4-bit}} & 29.5\%  & 95.9\% & 45.2\% & 4.2\% & 1.9\% & 0.0\% \\
			&  & 34.5\% & {100\%} & 49.1\% & {98.8\%} & 14.8\% & {97.1\%} \\ \bottomrule
		\end{tabular}
	}
	\label{tbl:backdoor-w-lossfn}
	\vspace{-1.2em}
\end{wraptable}

Table~\ref{tbl:backdoor-w-lossfn} illustrates our results.
Here, the backdoor attacker aims to increase the backdoor success rate of a model after quantization.
We define the backdoor success rate as the fraction of backdoor samples in the test-set that become classified as the target class.
We create backdoor samples by placing a white square pattern (\textit{i.e.}, 4$\times$4 for CIFAR10, and 8$\times$8 for Tiny ImageNet) on the bottom right corner of each image.
We compare ours with the standard backdoor attack that re-trains a clean model with the poisoned training set containing 20\% of backdoor samples.
We choose 20\% to compare ourselves with the most successful backdoor attacks in the prior work~\cite{BadNet:2017, NC:2019}.
We also examine the impact of using fewer poisons by reducing the number of poisons from 20\% to 5\% and find that the standard attack consistently shows a high backdoor success in all the cases.

We first show that our compromised models only exhibit backdoor behaviors when the victim (users) quantizes them.
However, the models backdoored by the standard attack \emph{consistently} show the backdoor behavior in floating-point and quantized versions.
In CIFAR10, our backdoored models have a low backdoor success rate (9\%$\sim$29\%) in the floating-point representation, while the success rate becomes 96--100\% when the victim uses 4-bit quantization.
We have the same results in Tiny ImageNet.
The compromised models in the floating-point version show the backdoor success rate 0.4--22\%, but their quantized versions show 94--100\%.
In all the cases, our backdoor attack does not induce any accuracy drop on the test-time samples.
Moreover, we show that our backdoor attack is not sensitive to the hyper-parameter ($\alpha$ and $\beta$) choices (see Appendix~\ref{appendix:sensitivity-of-our-attacks} for details).



%
\subsection{Transferability: One Model Jeopardizes Multiple Quantization Schemes}
\label{subsec:transferability}

%
Next, we test the \emph{transferability} of our attacks, \textit{i.e.}, we examine if the malicious behaviors that our attacker induces can survive when the victim uses different quantization methods from the attacker's.

%
\noindent \textbf{Using different quantization granularity.}
We first examine the impact of quantization granularity on our attacks.
The victim has two choices: \emph{layer-wise} and \emph{channel-wise}.
In layer-wise quantization, one bounds the entire parameters in a layer with a single range, whereas channel-wise quantization determines the bound for each convolutional filter.
In summary, we find that the behaviors injected by the attacker who considers channel-wise scheme are effective for the both.
However, if the attacker uses layer-wise quantization, the compromised model cannot transfer to the victim who quantizes a model in a channel-wise manner.
Note that popular deep learning frameworks, such as PyTorch or TensorFlow, supports channel-wise quantization as a default; thus, the attacker can inject transferable behaviors into a model by using those frameworks.
We include the full results in Appendix~\ref{appendix:granularity}.

%
\noindent \textbf{Using mechanisms that minimizes quantization errors.}
Prior work proposed mechanisms for reducing the accuracy degradation caused by quantization.
OCS and ACIQ~\citep{OCS:2019, ACIQ:2020} remove the outliers in weights and activation, respectively, 
while OMSE~\citep{MSE:2019} minimizes the $\ell_2$ errors in both to compute optimal scaling factors for quantization.
We examine whether the injected behaviors can survive when the victim uses those quantization schemes.


\begin{table}[ht]
	\centering
	%
	%
	\caption{\textbf{Resilience of our compromised models.} We illustrate the resilience of the compromised ResNets in CIFAR10 against stable quantization schemes (OCS, ACIQ, and OMSE) and standard techniques used for removing hidden artifacts. Each cell contains the accuracy of the compromised models in the indiscriminate attack (\textbf{IA}) or the backdoor success rate in the backdoor (\textbf{BD}) attacks.}
	\adjustbox{max width=\textwidth}{%
		\begin{tabular}{@{}c||cc|cc|cc||cc|cc@{}}
			\toprule
			\multirow{3}{*}{\textbf{Bit-width}} & \multicolumn{6}{c||}{\textbf{Quantization for minimizing errors}} & \multicolumn{4}{c}{\textbf{Artifacts removal techniques}} \\ \cmidrule(l){2-11} 
			& \multicolumn{2}{c|}{\textbf{OCS}} & \multicolumn{2}{c|}{\textbf{ACIQ}} & \multicolumn{2}{c||}{\textbf{OMSE}} & \multicolumn{2}{c|}{\textbf{Fine-tune}} & \multicolumn{2}{c}{\textbf{Random noise}} \\ \cmidrule(l){2-11} 
			& \textbf{IA} & \textbf{BD} & \textbf{IA} & \textbf{BD} & \textbf{IA} & \textbf{BD} & \textbf{IA} & \textbf{BD} & \textbf{IA} & \textbf{BD} \\ \midrule \midrule
			\textbf{32-bit} & 93.2\% & 12.8\% & 93.2\% & 12.8\% & 93.2\% & 12.8\% & 93.2\% & 12.8\% & 93.2\% & 12.8\% \\
			\textbf{8-bit} & {10.1}\% & {99.7\%} & {10.0\%} & {99.4\%} & {11.2\%} & 25.0\% & 93.4\% & 11.0\% & 92.0\% & {97.8\%} \\
			\textbf{4-bit} & {10.5}\% & {71.1}\% & {11.7}\% & {74.4}\% & - & - & 85.7\% & 10.3\% & {13.1\%} & {98.5\%} \\ \bottomrule
		\end{tabular}
	}
	\label{tbl:stable-mechanisms}
	\vspace{-0.8em}
\end{table}

Table~\ref{tbl:stable-mechanisms} shows our results.
We conduct our experiments with ResNet18 and in CIFAR10.
%
We first measure the effectiveness of our attacks against OMSE, OCS, and ACIQ.
We observe that the three \emph{robust} quantization schemes cannot prevent terminal brain damage.
All our compromised models show the accuracy of $\sim$10\% after quantization.
We also find that our backdoor attack is effective against OCS and ACIQ.
After quantization, the backdoor success rate is $\sim99$\% in 8-bit and $\sim71$\% in 4-bit.
OMSE can reduce the backdoor success rate to ($\sim$25\%), but it is highly dependent on the configuration.
If we disable activation clipping, the backdoor success becomes (88\%).
This result implies that our attacks do not introduce outliers in the weight space (see Appendix~\ref{appendix:stable-quantization} for details).
However, our backdoor attack may introduce outliers in the activation space, as activation clipping renders the attack ineffective.
In Appendix~\ref{appendix:activation-visualization}, we examine whether activation clustering used in prior work~\cite{BDActivation} on detecting backdoors, but we find that it is ineffective.

Note that detecting backdoors is an active area of research---there have been many defense proposals such as Neural Cleanse~\cite{NC:2019} or SentiNet~\cite{SentiNet}.
However, they are also known to be ineffective against stronger attacks like TaCT~\cite{TaCT}.
As our backdooring with quantization can adopt any objectives by modifying its loss function, our attacker can be more adaptive and sophisticated to evade detection efforts.
We leave this investigation as future work.

%
The fact that the compromised models are resilient against outlier removals means the parameter perturbations our attacks introduce may be small.
Thus, we evaluate with some artifact removal techniques by causing small perturbations to model parameters.
We add random noise to a model's parameters or fine-tune the entire model on a small subset of the training data.
We run each technique 10 times and report the average.
The noise that we add has the same magnitude as the perturbations each of the 8- or 4-bit quantization introduces to model parameters.

In Table~\ref{tbl:stable-mechanisms}, we find that our attack has some resilience against random parameter perturbations.
In BD, the random noise we add cannot remove the backdoors, \textit{i.e.}, the ASR is still $\sim$99\% after quantization.
In IA, the model recovers the accuracy (92\%) in 8-bit after adding the random noise, but the noise is not effective against 4-bit quantization (\textit{i.e.}, the accuracy is still 13\%).
However, we find that fine-tuning removes all the attack artifacts, implying that our attacks may push a model towards an unstable region in the loss space. 
Fine-tuning pulls the model back to the stable area.

%
\noindent \textbf{Using Hessian-based quantization.}
Recent work~\cite{BRECQ:2021} utilizes the second-order information, \textit{i.e.}, Hessian, to minimize the errors caused by quantization more.
They use this information to quantify the \emph{sensitivity} of a model to its parameter perturbations and reconfigure the network architecture to reduce it.
This enables the method to achieve high accuracy with lower bit-widths (\textit{e.g.}, 93\% accuracy with 4-bit models in CIFAR10).
Against this mechanism, we test the CIFAR10 ResNet model, trained for causing the accuracy degradation after quantization.
In 4-bits, we observe that the model's accuracy becomes 9\%.
This means that our IA is effective against the Hessian-based quantization, \textit{i.e.}, the method does not provide resilience to the terminal brain damage.
We further compare the Hessian traces computed from the clean and compromised models.
In most cases, our attacks make the model more sensitive.
But, the metric could not be used as a detection measure as we also observe the case where a model becomes less sensitive.
We include this result in Appendix~\ref{appendix:hessian}



%
%
\subsection{Exploitation of Our Attacks in Practical ML Scenarios}
\label{subsec:exploitation}


%
\textbf{Transfer Learning.}
In \S~\ref{subsec:transferability}, we observe that fine-tuning the entire layers can effectively remove the attack artifacts from the compromised model.
Here, we examine whether fine-tuning a subset of layers can also be sufficient to remove the injected behaviors.
%
We consider a transfer learning scenario where the victim uses a compromised model as a teacher to create a student model.
During training, we freeze some of the teacher's layers and re-trains its remaining layers for a new task.
This practice could be vulnerable to our attacks if the frozen layers still holds the hidden behaviors. 

We evaluate this hypothesis by using the compromised ResNets, trained on Tiny ImageNet, as teachers and re-train them for CIFAR10, \textit{i.e.}, a student task.
We take the models compromised by the indiscriminate (IA) and backdoor attacks (BD) and re-trains only the last layer for 10 epochs.
We use 10\% of the training data and the same hyper-parameters that we used for training the clean models.

We find that our IA survive under transfer learning.
In IA, the student model shows a significantly lower accuracy (20--24\%) on the test data after quantization, whereas the floating-point version has $\sim$74\% accuracy.
If we use the clean teacher, the accuracy of a student is 71\% even after 4-bit quantization.
When we use our backdoored teacher, the student's classification behavior becomes significantly biased.
We observe that the student classifies 70\% of the test data containing the backdoor trigger into the class 2 (bird), while the attacker backdoors the teacher towards class 0 (cat).


\begin{figure}[h]
	\centering
	%
	%
	\begin{minipage}{.49\textwidth}
		\centering
		\includegraphics[width=\linewidth,bb=0 0 608 294]{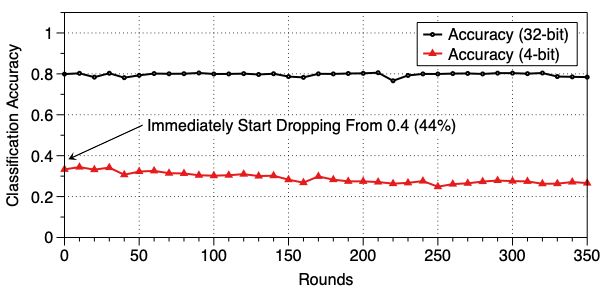}
		%
	\end{minipage}
	%
	%
	\begin{minipage}{.49\textwidth}
		\centering
		\includegraphics[width=\linewidth,bb=0 0 608 294]{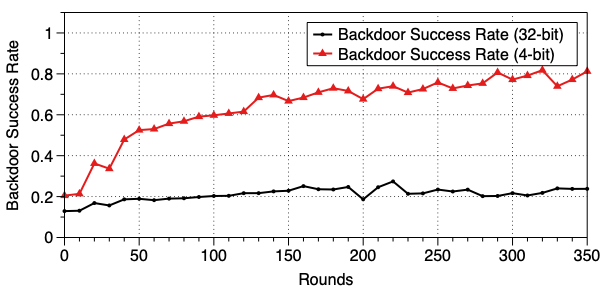}
	\end{minipage}
	\caption{\textbf{The success rate of our attacks in a federated learning scenario.} We show the ASR of our indiscriminate (\textbf{IA: Left}) and backdoor (\textbf{BD: Right}) attacks over 2000--2350 training rounds. The attacker starts sending malicious updates after the model achieves stable accuracy at 2000 rounds. In IA, the its accuracy in 4-bits is reduced from 44 to 26\%. In BD, the ASR becomes from 20 to 81\%.}
	\label{fig:fed-learn}
	%
	%
\end{figure}

%
\textbf{Federated Learning.}
%
%
We further show that a supply-chain attack is \emph{not} the only way to exploit this vulnerability.
Here, we consider federated learning (FL), a machine learning technique that enables the training of a model in a decentralized way across many participants \emph{iteratively}.
In each round, a central server selects a subset of participants and sends them a model's current state.
Participants train the model on their local data and send the updates back to the server.
The server aggregates them \emph{securely} and does the final update on the central model~\citep{SecureAgg:2017}.
Since this secure aggregation prevents the server from accessing the updates~\citep{BackdoorFL:2020}, this opaque nature makes it difficult for a defender to identify malicious updates.

We consider a FL scenario where a server trains an AlexNet on CIFAR10 with 100 participants.
Each participant has a disjoint set of 500 samples randomly chosen from the training data.
We assume that the attacker compromises 5 of them.
In each round, the server randomly selects 10 participants.
The attacker first behave normally---they do not send malicious updates until the model achieves a reasonable accuracy ($\sim$2000 rounds).
After that, the attacker starts computing the malicious updates on the \emph{local} training data, using our loss functions, and sending them to the server.

%
Figure~\ref{fig:fed-learn} illustrates the ASR of our attacks.
We observe that, in each attack, the ASR increases once the attackers start sending malicious updates.
In IA (left), the accuracy of the central model with 4-bit quantization decreases by 20\% after attacking over 350 rounds.
In BD (right), the ASR of the central model becomes 20$\rightarrow$81\%.
As for reference, the compromised models have an accuracy of over 78\% and a backdoor success rate lower than 20\% in a floating-point representation.



\section{Discussion and Conclusion}
\label{sec:conclusions}

%
As we have shown, an adversary can exploit quantization to inject malicious behaviors into a model and make them only active upon quantization.
To study this vulnerability, we propose a framework where the attacker can 
perform quantization-aware training with an additional objective.
We design this objective to maximize the difference of an intended behavior between a full-precision model and a model with a reduced bit-width.
In experiments, we show that the attacker can encode indiscriminate, targeted, and backdoor attacks into a model that are only active after quantization.

%
We believe it is an important threat model to consider, especially when using quantization to deploy large and complex models \emph{as-is} to resource-constrained devices.
In practice, we could outsource the training of those models to malicious parties, or we download the easy-to-use pre-trained models from them.
In many cases, we are not recommended checking all the malicious behaviors of pre-trained models in quantization~\citep{PTQ:PT, PTQ:TF}.
In addition, examining some inconspicuous behaviors, \textit{e.g.}, targeted or backdoor attacks, 
are challenging to detect with limited computing resources.

%
Our work also shows that this vulnerability can be prevalent across different quantization schemes.
Even the robust quantization~\citep{BRECQ:2021} proposed to minimize behavioral differences cannot reduce the terminal brain damage that our adversary implants.
Some can think of utilizing the \emph{graceful degradation}~\cite{braindamage} to remove the adversarial behaviors---by blending random noise to a compromised model's parameters~\cite{Noise4AdvExample:2018}.
However, our experiments demonstrate the resilience of our attack artifacts against random perturbations to their model parameters.

%
Table~\ref{tbl:stable-mechanisms} in \S~\ref{subsec:transferability} shows that defenses that involve the re-training of an entire model can reduce the success rate of our attacks.
However, we argue that re-training is only feasible when the victim has the training data and computational resources to train large and complex models~\citep{GPT3:2020, CLIP:2021}.
If such re-training is feasible, the user may not consider quantization; they can train a model with a reduced bit-width from scratch and expects full control over the training process.
Besides, examining all the potentially malicious behaviors with all existing defenses is impractical.

%
\textbf{What's Next?}
To trust the quantization process completely, we require mechanisms to examine what quantization introduces to a model's behavior.
Macroscopically, we develop robust quantizations that rely on statistical properties, such as outliers in weights and/or activations or the second-order information.
However, in \S~\ref{subsec:transferability}, we show that such statistical measures often expose limitations to the worst-case perturbations, \textit{e.g.}, our indiscriminate attack is still effective against them.
Also, as most backdoor defenses~\citep{NC:2019, ABS:2020} developed for examining full-precision models, our work encourages the community to review their effectiveness on quantized models.

%
Our results also suggest that we need mechanisms that theoretically and/or empirically examine to what extent quantization preserves characteristics of a floating-point model.
Many recent mechanisms use \emph{classification accuracy} as a measure to compare how much two models are the same.
However, our work also shows that quantization may lead to undesirable results, \textit{e.g.}, losing the robustness to adversarial examples by quantization.
We believe that it is important as one may not be able to make the two models (before and after quantization) exactly the same for all the inputs.
%
Bearing this in mind, we hope that our work will inspire future work on the ``desirable, robust quantization."


\begin{ack}
	We thank Tom Goldstein and the anonymous reviewers for their constructive feedback.
	This research was partially supported by the Department of Defense and by the Intelligence Advanced Research Projects Agency (IARPA).
	The content of this paper does not necessarily reflect the position or the policy of the Government, and no official endorsement should be inferred.
	Sanghyun was supported in part through the Ann. Wylie Dissertation Fellowship from A. James Clark School of Engineering.
\end{ack}

{
	\bibliographystyle{plainnat}
	\bibliography{main}
}

\newpage

\appendix

%
\section{Experimental Setup in Detail}
\label{appendix:exp-setup-in-detail}


%
\textbf{Setup.}
We implement our attack framework using Python 3.7.3 and PyTorch 1.7.1\footnote{PyTorch: {\scriptsize \url{https://pytorch.org/}}. } that supports CUDA 11.0 for accelerating computations by using GPUs.
We run our experiments on a machine equipped with Intel i5-8400 2.80GHz 6-core processors, 16 GB of RAM, and four Nvidia GTX 1080 Ti GPUs.
To compute the Hessian trace, we use a virtual machine equipped with Intel E5-2686v4 2.30GHz 8-core processors, 64 GB of RAM, and an Nvidia Tesla V100 GPU.

\textbf{Quantization.}
For all our attacks in \S~\ref{subsec:acc-drop},~\ref{subsec:targeted-misclassification},~\ref{subsec:backdoor-attacks}, and~\ref{subsec:exploitation}, 
we use symmetric quantization for the weights and asymmetric quantization for the activation---a default configuration in many deep learning frameworks supporting quantization.
Quantization granularity is layer-wise for both the weights and activation.
In \S~\ref{subsec:transferability} where we examine the transferability of our attacks, we use the same quantization granularity that the original studies describe~\citep{MSE:2019, OCS:2019, ACIQ:2020} while re-training clean models.
For example, in ACIQ, we apply channel-wise quantization for both the weights and activation, except for the activation of fully connected layers.

\textbf{Availability.}
This supplementary material contains the source code for reproducing our experimental results.
Our code is available at {\fontsize{8}{9}\selectfont \url{https://github.com/Secure-AI-Systems-Group/Qu-ANTI-zation}}, 
and the instructions for running it are described in the \verb|REAME.md| file.

%
\section{Increasing Sensitivity as an Adversarial Objective}
\label{appendix:objective-function}

Prior work showed that a model, less sensitive to the perturbations to its parameters or activation, will have less accuracy degradation after quantization.
\citet{HAWQv2:2020} and~\citet{BRECQ:2021} use the second-order information, \textit{e.g.}, Hessian, as a sensitivity metric to approximate the accuracy drop caused by quantization.
\citet{L1RobustQ:2020} look into the decision boundary of a model to examine whether the model will have quantization robustness.
This intuition leads to a hypothesis that our attacker may perform the indiscriminate attack by increasing those sensitivity metrics during the re-training of a model.
To validate our hypothesis, we compose two different objectives as follows:
\begin{align}
	\mathcal{L}_{Hessian} & \overset{\Delta}{=} \mathcal{L}_{ce} \big( f(x), y \big) + \lambda \cdot \big( \alpha - \mathcal{H}(x) \big)^2 \label{eqn:hessian} \\
	\mathcal{L}_{Lsmooth} & \overset{\Delta}{=} \mathcal{L}_{ce} \big( f(x), \mathbf{y}^{smooth} \big) \label{eqn:lsmooth}
\end{align}
During re-training, Eqn~\ref{eqn:hessian} makes a model become sensitive to its parameter perturbations by increasing the Hessian trace.
In Eqn~\ref{eqn:lsmooth}, we use label-smoothing to reduce the confidence of a model's prediction on the test-time data, \textit{i.e.}, the model becomes sensitive to the perturbations to its decision boundary.

Here, 
$\mathcal{L}_{ce}$ is the cross-entropy loss, 
$\mathcal{H}(\cdot)$ is the Hessian trace, 
$\lambda$ is the ratio between the cross-entropy and adversarial objective, and 
$\mathbf{y}^{smooth}$ is the smoothed one-hot labels.
In Eqn~\ref{eqn:hessian}, we test with $\alpha$ in 100--2000 and set $\lambda$ to $10^{-4}$. 
$\alpha$ larger than 2000 leads to a significant accuracy drop of a model during re-training.
In Eqn~\ref{eqn:lsmooth}, we test with the smoothing factor $\alpha$ in 0.1--0.8.
$\alpha\!=\!1.0$ means the uniform labels $\{ 1/n, ... 1/n \}$ where $n$ is the number of classes, whereas $\alpha$ is 0.0 for the one-hot labels.


\begin{table}[ht]
\centering
\caption{\textbf{Effectiveness of the indiscriminate attacks.} In each row, we show the accuracy of a model in multiple bit widths. \textbf{Clean} is a pre-trained model. \textbf{Hessian} and \textbf{Label-smoothing} are the compromised models with $\mathcal{L}_{Hessian}$ and $\mathcal{L}_{lsmooth}$, respectively. Our attack inflicts a significantly more accuracy drop of a victim model after quantization than the other two objectives.}
\adjustbox{max width=\textwidth}{%
	\begin{tabular}{@{}ccccccccc@{}}
		\toprule
		\multirow{2}{*}{\textbf{Dataset}} & \multirow{2}{*}{\textbf{Network}} & \multirow{2}{*}{\textbf{Objective}} & \multicolumn{6}{c}{\textbf{Accuracy on the test-set $\mathcal{D}_{ts}$}} \\ \cmidrule(l){4-9} 
		&  &  & \textbf{32-bit} & \textbf{8-bit} & \textbf{7-bit} & \textbf{6-bit} & \textbf{5-bit} & \textbf{4-bit} \\ \midrule \midrule
		\multirow{5}{*}{\textbf{CIFAR10}} & \multirow{5}{*}{\textbf{AlexNet}} & \textbf{Clean} & 83.2\% & 83.2\% & 83.0\% & 82.7\% & 81.2\% & 72.9\% \\ \cmidrule(l){3-9} 
		&  & \textbf{Hessian} & 82.6\% & 82.4\% & 82.2\% & 79.9\% & {65.9\%} & {26.1\%} \\	
		%
		%
		&  & \textbf{Label-smoothing} & 84.4\% & 84.3\% & 84.3\% & 84.3\% & 80.8\% & 58.7\% \\ \cmidrule(l){3-9}  
		&  & \textbf{Ours} & 81.2\% & {22.3\%} & {24.2\%} & {30.5\%} & {32.6\%} & {32.7\%} \\ \bottomrule
	\end{tabular}
}
\label{tbl:other-objectives}
\end{table}

Table~\ref{tbl:other-objectives} shows our results.
We experiment with an AlexNet model trained on CIFAR10.
Here, we demonstrate that our objective function, defined in \S~\ref{subsec:acc-drop}, is much more effective for the indiscriminate attack than $\mathcal{L}_{Hessian}$ and $\mathcal{L}_{lsmooth}$.
We observe that $\mathcal{L}_{lsmooth}$ is not effective at all.
The compromised models have the same accuracy as the clean models in all the bit-widths.
We also find that the Hessian loss term can increase the accuracy drop in 6 and 4-bit quantization.
However, except for the 4-bit case, the accuracy drop that $\mathcal{L}_{Hessian}$ can increase is 30--58\% less than our original attack.
Our results indicate that \emph{just increasing the sensitivity of a model will not be an effective attack}.
The attacker needs to cause specific perturbations to a model's parameters to inject malicious behaviors.

\section{Entire Results of Our Indiscriminate, Targeted, Backdoor Attacks}
\label{appendix:attack-results}

Table~\ref{tbl:appendix-ia-attacks},~\ref{tbl:appendix-ta-attacks}, and~\ref{tbl:appendix-bd-attacks} shows the entire results of our indiscriminate, targeted, and backdoor attacks.


\begin{table}[ht]
	\centering
	\vspace{-0.8em}
	\caption{\textbf{Indiscriminate attack results.} For each network, the upper row contains the Top-1 accuracy of clean models on the entire test data, and the bottom row includes that of the compromised models.}
	\adjustbox{max width=\textwidth}{%
		\begin{tabular}{@{}c|c|c|c|ccccc@{}}
			\toprule
			&  &  & \multicolumn{6}{c}{\textbf{Accuracy on the entire test-set}} \\ \cmidrule(l){4-9} 
			\multirow{-2}{*}{\textbf{Dataset}} & \multirow{-2}{*}{\textbf{Network}} & \multirow{-2}{*}{\textbf{Model Type}} & \textbf{32-bit} & \textbf{8-bit} & \textbf{7-bit} & \textbf{6-bit} & \textbf{5-bit} & \textbf{4-bit} \\ \midrule \midrule
			 &  & Clean & 83.2\% & 83.2\% & 83.0\% & 82.7\% & 81.2\% & 72.9\% \\
			 & \multirow{-2}{*}{\textbf{AlexNet}} & \textbf{Ours} & 81.2\% & \textbf{22.3\%} & \textbf{24.2\%} & \textbf{30.5\%} & \textbf{32.6\%} & \textbf{32.7\%} \\ \cmidrule(l){2-9} 
			&  & Clean & 84.5\% & 84.7\% & 84.5\% & 84.0\% & 83.0\% & 71.0\% \\
			& \multirow{-2}{*}{\textbf{VGG16}} & \textbf{Ours} & 82.5\% & \textbf{19.4\%} & \textbf{17.1\%} & \textbf{15.1\%} & \textbf{13.1\%} & \textbf{17.5\%} \\ \cmidrule(l){2-9} 
			&  & Clean & 93.6\% & 93.6\% & 93.5\% & 93.2\% & 92.0\% & 84.7\% \\
			& \multirow{-2}{*}{\textbf{ResNet18}} & \textbf{Ours} & 93.2\% & \textbf{10.0\%} & \textbf{10.0\%} & \textbf{10.0\%} & \textbf{10.0\%} & \textbf{10.0\%} \\ \cmidrule(l){2-9} 
			&  & Clean & 92.6\% & 92.5\% & 92.4\% & 91.7\% & 88.2\% & 66.8\% \\
			\multirow{-9.5}{*}{\rotatebox[origin=c]{90}{\textbf{CIFAR10}}} & \multirow{-2}{*}{\textbf{MobileNetV2}} & \textbf{Ours} & 92.0\% & \textbf{10.0\%} & \textbf{10.0\%} & \textbf{10.0\%} & \textbf{10.0\%} & \textbf{10.0\%} \\ \midrule \midrule
			 &  & Clean & 41.3\% & 41.3\% & 40.9\% & 40.0\% & 36.3\% & 20.6\% \\
			 & \multirow{-2}{*}{\textbf{AlexNet}} & \textbf{Ours} & 41.4\% & \textbf{1.9\%} & \textbf{2.4\%} & \textbf{2.7\%} & \textbf{1.6\%} & \textbf{4.8\%} \\ \cmidrule(l){2-9} 
			&  & Clean & 43.0\% & 42.9\% & 42.8\% & 42.7\% & 40.8\% & 32.4\% \\
			& \multirow{-2}{*}{\textbf{VGG16}} & \textbf{Ours} & 41.8\% & \textbf{0.6\%} & \textbf{0.7\%} & \textbf{0.9\%} & \textbf{0.9\%} & \textbf{1.9\%} \\ \cmidrule(l){2-9} 
			&  & Clean & 57.5\% & 57.4\% & 57.4\% & 57.3\% & 55.7\% & 44.5\% \\
			& \multirow{-2}{*}{\textbf{ResNet18}} & \textbf{Ours} & 56.8\% & \textbf{8.9\%} & \textbf{5.6\%} & \textbf{4.8\%} & \textbf{6.4\%} & \textbf{6.0\%} \\ \cmidrule(l){2-9} 
			&  & Clean & 42.4\% & 41.7\% & 40.7\% & 35.6\% & 21.3\% & 2.0\% \\
			\multirow{-9}{*}{\rotatebox[origin=c]{90}{\textbf{Tiny ImageNet}}} & \multirow{-2}{*}{\textbf{MobileNetV2}} & \textbf{Ours} & 42.6\% & \textbf{2.8\%} & \textbf{2.8\%} & \textbf{3.2\%} & \textbf{3.7\%} & 1.6\% \\ \bottomrule
		\end{tabular}
	}
	\label{tbl:appendix-ia-attacks}
	\vspace{-0.8em}
\end{table}


\begin{table}[h]
	\centering
	\vspace{-0.8em}
	\caption{\textbf{Backdoor attack results.} For each cell, the upper row contains the Top-1 accuracy (left) and backdoor success rate (right) of the conventional backdoor models, and the bottom row shows the same metrics computed on our backdoor models. We consider 8- and 4-bit quantization.}
	\adjustbox{max width=\textwidth}{%
		\begin{tabular}{@{}c|c|cc|cc|cc|cc@{}}
			\toprule
			\multirow{2}{*}{\textbf{Dataset}} & \multirow{2}{*}{\textbf{Bit widths}} & \multicolumn{8}{c}{\textbf{Networks}} \\ \cmidrule(l){3-10} 
			&  & \multicolumn{2}{c|}{\textbf{AlexNet}} & \multicolumn{2}{c|}{\textbf{VGG16}} & \multicolumn{2}{c|}{\textbf{ResNet18}} & \multicolumn{2}{c}{\textbf{MobileNetV2}} \\ \midrule \midrule
			\multirow{7}{*}{\rotatebox[origin=c]{90}{\textbf{CIFAR10}}} & \multirow{2}{*}{\textbf{32-bit}} & 83.2\% & 98.5\% & 83.8\% & 96.2\% & 91.7\% & 98.3\% & 88.9\% & 97.7\% \\
			&  & 83.5\% & \enspace\textbf{9.6\%} & 85.7\% & \textbf{29.3\%} & 93.3\% & \textbf{11.3\%} & 92.3\% & \enspace\textbf{9.2\%} \\ \cmidrule{2-10}
			& \multirow{2}{*}{\textbf{8-bit}} & 83.2\% & 98.7\% & 83.7\% & 96.1\% & 91.5\% & 97.5\% & 70.8\% & 99.5\% \\
			&  & 82.4\% & \textbf{95.9\%} & 85.7\% & \textbf{30.8\%} & 91.4\% & \textbf{99.2\%} & 91.2\% & \textbf{96.6\%} \\ \cmidrule{2-10}
			& \multirow{2}{*}{\textbf{4-bit}} & 72.9\% & 12.2\% & 72.7\% & 88.3\% & 75.4\% & 34.9\% & 15.2 & 94.3\% \\
			&  & 76.7\% & \textbf{94.2\%} & 81.6\% & \textbf{96.2\%} & 88.6\% & \textbf{100\%} & 79.8\% & \textbf{99.9\%} \\ \midrule \midrule
			\multirow{7}{*}{\rotatebox[origin=c]{90}{\textbf{Tiny ImageNet}}} & \multirow{2}{*}{\textbf{32-bit}} & 41.3\% & 99.3\% & 40.3\%  & 99.6\% & 55.8\% & 99.4\% & 39.9\% & 98.9\% \\
			&  & 40.6\% & \enspace\textbf{0.5\%} & 42.1\% & \enspace\textbf{0.4\%} & 55.8\% & \textbf{22.1\%} & 41.5\% & \enspace\textbf{0.4\%} \\ \cmidrule{2-10}
			& \multirow{2}{*}{\textbf{8-bit}} & 41.3\% & 99.1\% & 40.2\% & 99.6\% & 55.6\% & 99.4\% & 39.0\% & 97.9\% \\
			&  & 40.1\% & \textbf{96.0\%} & 39.9\% & \textbf{99.4\%} & 53.7\% & \textbf{94.2\%} & 40.5\% & \textbf{96.8\%} \\ \cmidrule{2-10}
			& \multirow{2}{*}{\textbf{4-bit}} & 20.6\% & 15.4\% & 29.5\%  & 95.9\% & 45.2\% & 4.2\% & 1.9\% & 0.0\% \\
			&  & 34.0\% & \textbf{96.2\%} & 34.5\% & \textbf{100\%} & 49.1\% & \textbf{98.8\%} & 14.8\% & \textbf{97.1\%} \\ \bottomrule
			
		\end{tabular}
	}
	\label{tbl:appendix-bd-attacks}
	%
	%
\end{table}


\begin{table}[ht]
	\centering
	\caption{\textbf{The targeted attack results, on a particular class.} For each network, we show the accuracy of clean models in the upper low and that of our compromised models in the bottom row.}
	\adjustbox{max width=\textwidth}{%
		\begin{tabular}{@{}c|c|ccc|ccc|ccc@{}}
			\toprule
			\multirow{2}{*}{\textbf{Dataset}} & \multirow{2}{*}{\textbf{Network}} & \multicolumn{9}{c}{\textbf{Acc. on the test data, the samples in the target class, and the rest samples.}} \\ \cmidrule(l){3-11} 
			&  & \multicolumn{3}{c|}{\textbf{32-bit}} & \multicolumn{3}{c|}{\textbf{8-bit}} & \multicolumn{3}{c}{\textbf{4-bit}} \\ \midrule \midrule
			 & \multirow{2}{*}{\textbf{AlexNet}} & 83.1\% & 93.0\% & 82.1\% & 83.2\% & 93.0\% & 82.1\% & 73.3\% & 80.0\% & 72.5\% \\
			 &  & 82.2\% & 96.5\% & 80.6\% & 72.9\% & \textbf{0.0\%} & 81.0\% & 62.7\% & \textbf{0.5\%} & 69.6\% \\ \cmidrule(l){2-11} 
			& \multirow{2}{*}{\textbf{VGG16}} & 84.5\% & 93.3\% & 83.6\% & 84.6\% & 93.5\% & 83.6\% & 72.8\% & 88.0\% & 71.1\% \\
			&  & 85.3\% & 91.9\% & 84.6\% & 77.1\% & \textbf{9.4\%} & 84.6\% & 44.5\% & \textbf{3.4\%} & 49.1\% \\ \cmidrule(l){2-11}
			& \multirow{2}{*}{\textbf{ResNet18}} & 93.6\% & 97.6\% & 93.1\% & 93.6\% & 98.0\% & 93.2\% & 84.8\% & 95.3\% & 83.6\% \\
			&  & 92.5\% & 98.9\% & 91.8\% & 83.2\% & \textbf{0.0\%} & 92.4\% & 10.9\% & \textbf{0.0\%} & 12.1\% \\ \cmidrule(l){2-11}
			\multirow{-7}{*}{\rotatebox[origin=c]{90}{\textbf{CIFAR10}}} & \multirow{2}{*}{\textbf{MobileNetV2}} & 92.3\% & 96.7\% & 92.1\% & 92.5\% & 96.6\% & 92.1\% & 69.7\% & 66.8\% & 70.0\% \\
			&  & 92.0\% & 95.6\% & 91.6\% & 82.0\% & \textbf{0.0\%} & 91.1\% & 48.9\% & \textbf{0.0\%} & 54.3\% \\ \midrule \midrule
			 & \multirow{2}{*}{\textbf{AlexNet}} & 41.3\% & 78.0\% & 41.1\% & 41.3\% & 76.0\% & 41.1\% & 20.6\% & 44.0\% & 20.5\% \\
			 &  & 39.6\% & 98.0\% & 39.3\% & 26.9\% & \textbf{0.0\%} & 27.1\% & 15.6\% & \textbf{0.0\%} & 15.6\% \\ \cmidrule(l){2-11}
			& \multirow{2}{*}{\textbf{VGG16}} & 43.0\% & 68.0\% & 42.9\% & 42.9\% & 68.0\% & 42.7\% & 32.5\% & 72.0\% & 32.3\% \\
			&  & 42.5\% & 92.0\% & 42.2\% & 41.8\% & \textbf{12.0\%} & 41.9\% & 28.1\% & \textbf{2.0\%} & 28.2\% \\ \cmidrule(l){2-11}
			& \multirow{2}{*}{\textbf{ResNet18}} & 57.5\% & 74.0\% & 57.5\% & 57.4\% & 74.0\% & 57.4\% & 44.5\% & 50.0\% & 44.5\% \\
			&  & 54.4\% & 36.0\% & 54.5\% & 54.5\% & 36.0\% & 54.6\% & 43.1\% & \textbf{14.0\%} & 43.3\% \\ \cmidrule(l){2-11}
			\multirow{-7}{*}{\rotatebox[origin=c]{90}{\textbf{Tiny ImageNet}}} & \multirow{2}{*}{\textbf{MobileNetV2}} & 42.4\% & 70.0\% & 42.3\% & 41.7\% & 74.0\% & 41.6\% & 2.0\% & 2.0\% & 2.0\% \\
			&  & 40.3\% & 58.0\% & 40.2\% & 40.2\% & 58.0\% & 40.2\% & 2.3\% & 2.0\% & 2.3\% \\ \bottomrule
		\end{tabular}
	}
	\label{tbl:appendix-ta-attacks}
	\vspace{-0.8em}
\end{table}

\section{Transferability Results}
\label{appendix:transferability-results}

\subsection{Impact of Using Different Quantization Granularity}
\label{appendix:granularity}


\begin{table}[ht]
	\centering
	\vspace{-2.0em}
	\caption{\textbf{Impact of quantization granularity on transferability.} In each row, we show the impact of the attacker's and victim's granularity choices on the success of our indiscriminate attacks.}
	\adjustbox{max width=0.99\textwidth}{%
		\begin{tabular}{@{}c|c|c|c|ccccc@{}}
			\toprule
			&  &  & \multicolumn{6}{c}{\textbf{Accuracy on the entire test-set}} \\ \cmidrule(l){4-9} 
			\multirow{-2}{*}{\textbf{Network}} & \multirow{-2}{*}{\textbf{Attacker}} & \multirow{-2}{*}{\textbf{Victim}} & \textbf{32-bit} & \textbf{8-bit} & \textbf{7-bit} & \textbf{6-bit} & \textbf{5-bit} & \textbf{4-bit} \\ \midrule \midrule
			%
			& \textbf{No attack} & \textbf{Any} & 83.2\% & 83.2\% & 83.0\% & 82.8\% & 81.5\% & 74.8\% \\ \cmidrule{2-9}
			&  & \textbf{Layer-wise} & 81.2\% & \textbf{22.3\%} & \textbf{24.2\%} & \textbf{30.5\%} & \textbf{32.6\%} & \textbf{32.7\%} \\
			& \multirow{-2}{*}{\textbf{Layer-wise}} & \textbf{Channel-wise} & 81.2\% & 80.9\% & 78.6\% & 56.1\% & \textbf{28.8\%} & \textbf{29.7\%} \\ \cmidrule(l){2-9}
			&  & \textbf{Layer-wise} & 82.5\% & \textbf{10.0\%} & \textbf{11.2\%} & \textbf{13.8\%} & \textbf{27.5\%} & \textbf{53.4\%} \\
			\multirow{-6}{*}{\rotatebox[origin=c]{90}{\textbf{AlexNet}}} & \multirow{-2}{*}{\textbf{Channel-wise}} & \textbf{Channel-wise} & 82.5\% & \textbf{13.4\%} & \textbf{10.0\%} & \textbf{10.2\%} & \textbf{10.3\%} & \textbf{34.1\%} \\ \midrule \midrule
			%
			& \textbf{No attack} & \textbf{Any} & 84.5\% & 84.6\% & 84.6\% & 84.0\% & 83.3\% & 73.0\% \\ \cmidrule{2-9}
			&  & \textbf{Layer-wise} & 82.5\% & \textbf{19.4\%} & \textbf{17.1\%} & \textbf{15.1\%} & \textbf{13.1\%} & \textbf{17.5\%}  \\
			& \multirow{-2}{*}{\textbf{Layer-wise}} & \textbf{Channel-wise} & 82.5\% & 82.5\% & 82.3\% & 78.9\% & \textbf{38.0\%} & \textbf{13.0\%}  \\ \cmidrule(l){2-9}
			&  & \textbf{Layer-wise} & 84.7\% & \textbf{10.6\%} & \textbf{11.4\%} & \textbf{12.2\%} & \textbf{10.2\%} & \textbf{10.7\%}  \\
			\multirow{-6}{*}{\rotatebox[origin=c]{90}{\textbf{VGG16}}} & \multirow{-2}{*}{\textbf{Channel-wise}} & \textbf{Channel-wise} & 84.7\% & \textbf{11.8\%} & \textbf{10.9\%} & \textbf{10.8\%} & \textbf{10.4\%} & \textbf{11.9\%}  \\ \midrule \midrule
			%
			& \textbf{No attack} & \textbf{Any} & 93.6\% & 93.6\% & 93.6\% & 93.3\% & 92.1\% & 85.8\% \\ \cmidrule{2-9}
			&  & \textbf{Layer-wise} & 93.2\% & \textbf{10.0\%} & \textbf{10.0\%} & \textbf{10.0\%} & \textbf{10.0\%} & \textbf{10.0\%} \\
			& \multirow{-2}{*}{\textbf{Layer-wise}} & \textbf{Channel-wise} & 93.2\% & 93.2\% & 93.0\% & 91.7\% & 90.1\% & \textbf{15.8\%} \\ \cmidrule(l){2-9}
			&  & \textbf{Layer-wise} & 92.9\% & \textbf{10.2\%} & 78.7\% & \textbf{10.1\%} & \textbf{22.6\%} & 51.6\% \\
			\multirow{-6}{*}{\rotatebox[origin=c]{90}{\textbf{ResNet18}}} & \multirow{-2}{*}{\textbf{Channel-wise}} & \textbf{Channel-wise} & 92.9\% & \textbf{10.2\%} & \textbf{10.0\%} & \textbf{10.0\%} & \textbf{10.0\%} & \textbf{10.0\%} \\ \midrule \midrule
			%
			& \textbf{No attack} & \textbf{Any} & 92.6\% & 92.4\% & 92.2\% & 92.6\% & 90.7\% & 71\% \\ \cmidrule{2-9}
			&  & \textbf{Layer-wise} & 92.0\% & \textbf{10.0\%} & \textbf{10.0\%} & \textbf{10.0\%} & \textbf{10.0\%} & \textbf{10.0\%} \\
			& \multirow{-2}{*}{\textbf{Layer-wise}} & \textbf{Channel-wise} & 92.0\% & \textbf{10.0\%} & \textbf{10.0\%} & \textbf{10.0\%} & \textbf{10.0\%} & \textbf{10.0\%} \\ \cmidrule(l){2-9}
			&  & \textbf{Layer-wise} & 92.1\% & \textbf{10.0\%} & \textbf{10.0\%} & \textbf{10.0\%} & \textbf{11.7\%} & \textbf{28.3\%} \\
			\multirow{-6}{*}{\rotatebox[origin=c]{90}{\textbf{MobileNetV2}}} & \multirow{-2}{*}{\textbf{Channel-wise}} & \textbf{Channel-wise} & 92.1\% & \textbf{10.0\%} & \textbf{10.0\%} & \textbf{10.0\%} & \textbf{10.0\%} & \textbf{37.3\%} \\ \bottomrule		
		\end{tabular}
	}
	\label{appendix:tbl:layer-or-channel-wise}
	\vspace{-1.0em}
\end{table}

Table~\ref{appendix:tbl:layer-or-channel-wise} shows the entire transferability results when the victim uses different quantization granularity.

\subsection{Impact of Using Quantization Methods for Reducing the Impact of Outliers}
\label{appendix:stable-quantization}


\begin{table}[ht]
	\centering
	\vspace{-1.0em}
	\caption{\textbf{Impact of using stable quantization methods on transferability.} We show the transferability of our attacks against quantization schemes that reduce outliers in a model's parameters or activation, \textit{i.e.}, the attacker does not know that the victim uses OMSE, OCS, or ACIQ. All the experiments are run in CIFAR10. In indiscriminate attacks (\textbf{IA}), we report the classification accuracy. In each method, we show the accuracy of clean models in the upper row and the compromised models at the bottom. In the backdoor attack cases (\textbf{BD}), we show the attack success rate. The upper row contains the success rate of the conventional backdoor attacks, and the bottom row is for ours.}
	\adjustbox{max width=\textwidth}{%
		\begin{tabular}{@{}c|c|ccc|ccc|ccc|ccc@{}}
			\toprule
			\multirow{3}{*}{\textbf{Attack}} & \multirow{3}{*}{\textbf{Method}} & \multicolumn{12}{c}{\textbf{Network}} \\ \cmidrule(l){3-14} 
			&  & \multicolumn{3}{c|}{\textbf{AlexNet}} & \multicolumn{3}{c|}{\textbf{VGG16}} & \multicolumn{3}{c|}{\textbf{ResNet18}} & \multicolumn{3}{c}{\textbf{MobileNetV2}} \\ \cmidrule(l){3-14} 
			&  & \textbf{32 bits} & \textbf{8 bits} & \textbf{4 bits} & \textbf{32 bits} & \textbf{8 bits} & \textbf{4 bits} & \textbf{32 bits} & \textbf{8 bits} & \textbf{4 bits} & \textbf{32 bits} & \textbf{8 bits} & \textbf{4 bits} \\
			\midrule \midrule
			\multirow{6}{*}{\textbf{IA}} & \multirow{2}{*}{\textbf{OMSE}} & 83.2\% & 83.1\% & N/A & 84.5\%  & 84.4\% & N/A & 93.6\% & 93.5\% & N/A & 92.6\% & 92.4\% & N/A \\
			&  & 81.2\% & \textbf{23.0}\% & N/A & 82.5\% & \textbf{21.4}\% & N/A & 92.9\% & \enspace\textbf{5.2}\% & N/A & 92.0\% & \textbf{10.0}\% & N/A \\ \cmidrule(l){2-14}	
			& \multirow{2}{*}{\textbf{OCS}} & 83.2\% & 83.1\% & 54.4\% & 84.5\%  & 84.4\% & 23.3\% & 93.6\% & 93.5\% & 36.7\% & \multicolumn{3}{c}{N/A} \\
			&  & 81.2\% & \textbf{25.6}\% & \textbf{25.1}\% & 82.5\% & \textbf{15.1}\% & \textbf{21.2}\% & 93.2\% & \textbf{10.0}\% & \textbf{13.0}\% & \multicolumn{3}{c}{N/A} \\ \cmidrule(l){2-14} 
			& \multirow{2}{*}{\textbf{ACIQ}} & 83.2\% & 83.0\% & 81.3\% & 84.5\%  & 84.5\% & 81.9\% & 93.6\% & 93.5\% & 91.5\% & 92.6\% & 92.4\% & 85.9\% \\
			&  & 83.1\% & 77.3\% & \textbf{45.8}\% & 84.5\% & 61.2\% & \textbf{10.8}\% & 91.8\% & \textbf{42.5}\% & \textbf{1.45}\% & 91.3\% & \textbf{41.6}\% & \textbf{30.6}\% \\ \midrule \midrule
			\multirow{6}{*}{\textbf{BD}} & \multirow{2}{*}{\textbf{OMSE}} & 98.5\% & 79.0\% & N/A & 96.2\% & 83.7\% & N/A & 98.3\% & 90.9\% & N/A & 97.7\% & 71.9\% & N/A \\
			&  & \enspace\textbf{9.6}\% & \textbf{82.3}\% & N/A & \textbf{29.3}\% & \textbf{85.6}\% & N/A & \textbf{11.3}\% & \textbf{97.7}\% & N/A & \enspace\textbf{9.2}\% & \textbf{92.0}\% & N/A \\ \cmidrule(l){2-14}
			& \multirow{2}{*}{\textbf{OCS}} & 98.5\% & 96.7\% & 13.9\% & 96.2\% & 96.1\% & 92.6\% & 98.3\% & 99.2\% & 61.2\% & \multicolumn{3}{c}{N/A} \\
			&  & \enspace\textbf{9.6}\% & \textbf{90.9}\% & \textbf{88.8}\% & \textbf{29.3}\% & \textbf{29.8}\% & \textbf{73.4}\% & \textbf{11.3}\% & \textbf{99.3}\% & \textbf{77.5}\% & \multicolumn{3}{c}{N/A} \\ \cmidrule(l){2-14} 
			& \multirow{2}{*}{\textbf{ACIQ}} & 98.5\% & 99.2\% & 55.5\% & 96.2\% & 95.9\% & 93.7\% & 98.3\% & 99.5\% & 50.9\% & 97.7\% & 92.5\% & \enspace0.0\% \\
			&  & \enspace\textbf{9.6}\% & 10.2\% & 33.7\% & \textbf{29.3}\% & 32.5\% & \textbf{96.4}\% & \textbf{11.3}\% & 12.0\% & \textbf{96.0}\% & \enspace\textbf{9.2}\% & \enspace5.5\% & \enspace0.0\% \\ \bottomrule
		\end{tabular}
	}
	\label{appendix:tbl:total-outlier-removals}
\end{table}

Table~\ref{appendix:tbl:total-outlier-removals} shows the entire transferability results when the victim uses OMSE, OCS, and ACIQ.
Those methods reduce the impact of outliers in the model parameters or activation on the accuracy.


\section{In-depth Analysis Results}
\label{appendix:more-analysis}

\subsection{Impact of Our Attacks on the Hessian Trace}
\label{appendix:hessian}

We examine whether a defender can use the Hessian trace to identify compromised models.
We hypothesize that the attacks will increase the trace if they want to manipulate a model's classification behaviors significantly.
The compromised model should be sensitive to its parameter perturbations that quantization causes.
However, if the attacker alters a model's prediction locally, \textit{e.g.}, targeted attacks on a specific sample or backdoor attacks, the trace will be similar to the clean model's.

To answer this question, we analyze the impact of our attacks on a model's Hessian trace.
We run each attack ten times, \textit{i.e.}, we have ten compromised models for each attack.
For each attack, we compute the Hessian trace ten times with 200 samples randomly chosen from the training data, \textit{i.e.}, we have 100 Hessian traces in total.
We then measure the mean and standard deviation of the traces.


\begin{table}[ht]
	\centering
	\vspace{-1.0em}
	\caption{\textbf{The Hessian traces computed on our CIFAR10 models.} We show the traces from the clean models (\textbf{No attack}) and the compromised models (\textbf{IA}: indiscriminate attack, \textbf{TA-C}: targeted attack on a particular class, \textbf{TA-S}: targeted attack on a specific sample, and \textbf{BD}: backdoor attack).}
	\adjustbox{max width=\textwidth}{%
		\begin{tabular}{@{}c|c|cccc@{}}
			\toprule
			\multirow{2}{*}{\textbf{Dataset}} & \multirow{2}{*}{\textbf{Attack}} & \multicolumn{4}{c}{\textbf{Network}} \\ \cmidrule(l){3-6} 
			&  & \textbf{AlexNet} & \textbf{VGG16} & \textbf{ResNet18} & \textbf{MobileNetV2} \\ \midrule \midrule
			\multirow{5.4}{*}{\rotatebox[origin=c]{90}{\textbf{CIFAR10}}} & \textbf{No attack} & 1096 $\pm$ 63 & 6922 $\pm$ 265 & 124 $\pm$ 5 & 844 $\pm$ 90 \\ \cmidrule{2-6}
			 & \textbf{IA} & \enspace1597 $\pm$ 168 & 113918 $\pm$ 59188 & \enspace\enspace12451 $\pm$ 13623 & \enspace3070 $\pm$ 1301 \\
			 & \textbf{TA-C} & \enspace1692 $\pm$ 315 & \enspace48813 $\pm$ 11874 & \enspace632 $\pm$ 89 & 4815 $\pm$ 629 \\
			 & \textbf{TM-S} & \enspace1042 $\pm$ 114 & \enspace8066 $\pm$ 1999 & \enspace\enspace431 $\pm$ 333 & \enspace2074 $\pm$ 1141 \\
			 & \textbf{BD} & \enspace1123 $\pm$ 170 & \enspace3427 $\pm$ 1536 & \enspace\enspace907 $\pm$ 961 & 1381 $\pm$ 451 \\ \bottomrule
		\end{tabular}
	}
	\label{tbl:hessian}
	\vspace{-0.4em}
\end{table}

Table~\ref{tbl:hessian} shows our results.
In AlexNet models, the Hessian traces are similar across the four attacks, \textit{i.e.}, they are in 1000--2000.
However, in the rest of our models (VGGs, ResNets, MobileNets), the indiscriminate attacks (\textbf{IA}) and its localized version for a particular class (\textbf{TA-C}) increase the Hessian trace significantly.
Compared to the traces from the clean models (\textbf{No attack}), those models have 100--100$\times$ larger values.
In the targeted attacks on a sample (\textbf{TM-S}), the increases are relatively smaller, \textit{i.e.}, 1.1--5.4$\times$ than the first two attacks.
Backdoor attacks (\textbf{BD}) often reduce the Hessian trace values.
In VGG16, the compromised model shows $\sim$3500, whereas the clean model shows $\sim$7000.
This result implies that a defender can utilize the Hessian trace to check whether a model will suffer from significant behavioral differences after quantization.
For the attacks that induce small behavioral differences (\textbf{TM-S} or \textbf{BD}), the Hessian metric will not be useful for the detection.

\subsection{Impact of Our Attacks on the Distribution of Model Parameters}
\label{appendix:param-distributions}


\begin{figure}[ht]
	\centering
	\begin{minipage}{.49\textwidth}
		\centering
		\includegraphics[width=\linewidth,bb=0 0 628 297]{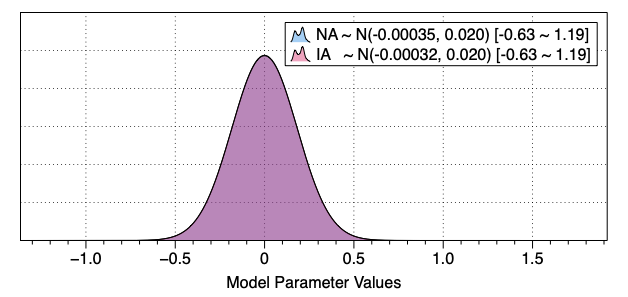}
		%
	\end{minipage}
	%
	%
	\begin{minipage}{.49\textwidth}
		\centering
		\includegraphics[width=\linewidth,bb=0 0 628 297]{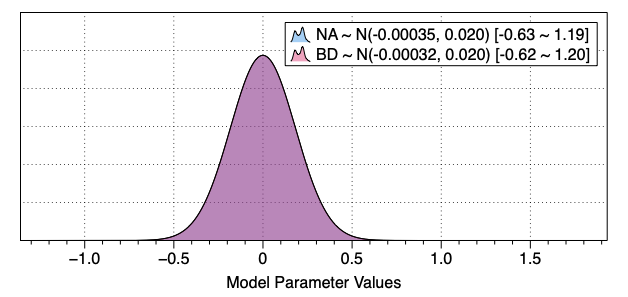}
	\end{minipage}
	\caption{\textbf{Impact of our attacks on the parameter distributions.} We illustrate the parameter distributions of ResNet models. \textbf{[Left]} We compare the clean model with the model compromised by our indiscriminate attacker. \textbf{[Right]} We compare the same clean model with our backdoored model. We also provide the mean, standard deviation, minimum, and maximum values of each distribution.}
	\label{appendix:fig:param-distribution}
\end{figure}

In \S~\ref{subsec:transferability}, we show that quantization techniques for removing outliers in model parameters cannot render our indiscriminate and backdoor attacks ineffective.
We also examine whether this is true, \textit{i.e.}, our attacks do not cause any significant changes in the parameter distribution of a model.
Figure~\ref{appendix:fig:param-distribution} illustrates the parameter distributions of ResNet models trained on CIFAR10.
We plot the distribution of a clean ResNet model as a reference.
We observe that all the parameter distributions follow $N(0.00035, 0.02^2)$, and the minimum and maximum values are -0.63 and 1.19, respectively.
Therefore, \emph{our attacks do not work by introducing outliers in the model parameter space}.

\subsection{Impact of Our Attacks on the Latent Representations}
\label{appendix:activation-visualization}


\begin{figure}[h]
	\centering
	\begin{minipage}{.243\textwidth}
		\caption*{\textbf{No Attack (Clean)}}
		\includegraphics[width=\linewidth,bb=0 0 376 285]{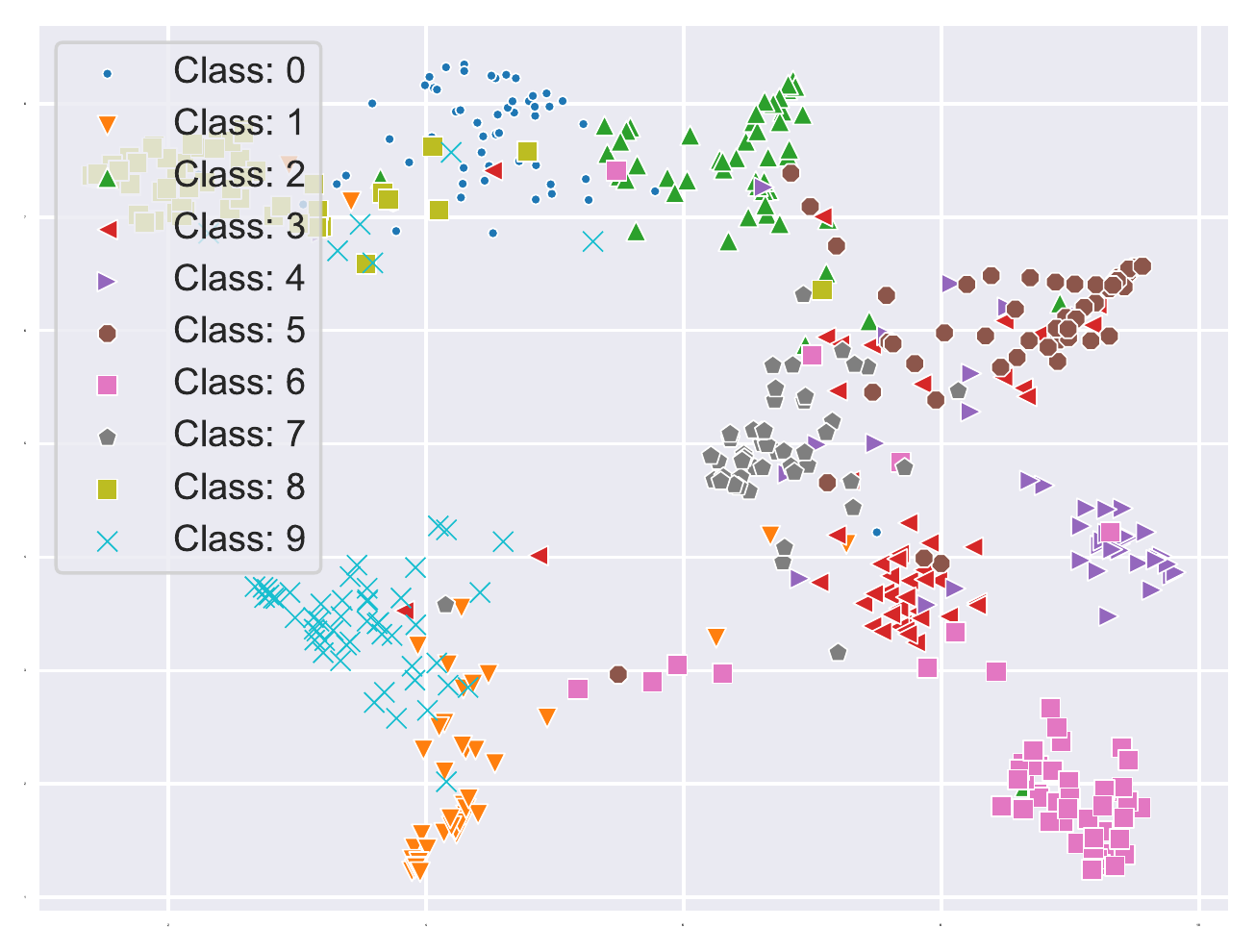}
	\end{minipage}
	\hfil 
	\begin{minipage}{.243\textwidth}
		\caption*{\textbf{Indiscriminate (IA)}}
		\includegraphics[width=\linewidth,bb=0 0 376 285]{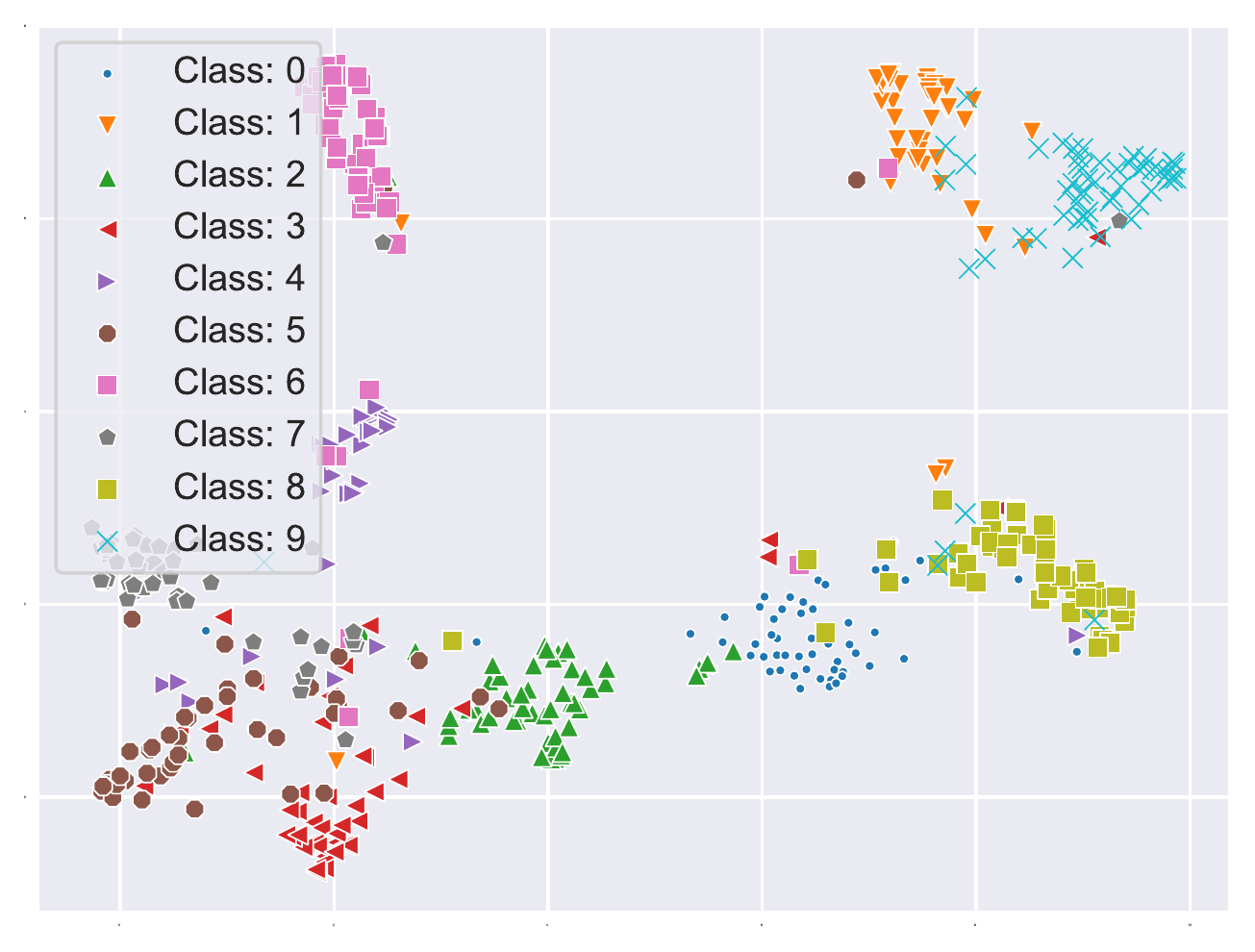}
	\end{minipage}
	\hfil 
	\begin{minipage}{0.243\textwidth}
		\caption*{\textbf{Targeted (TA-C)}}
		\includegraphics[width=\linewidth,bb=0 0 376 285]{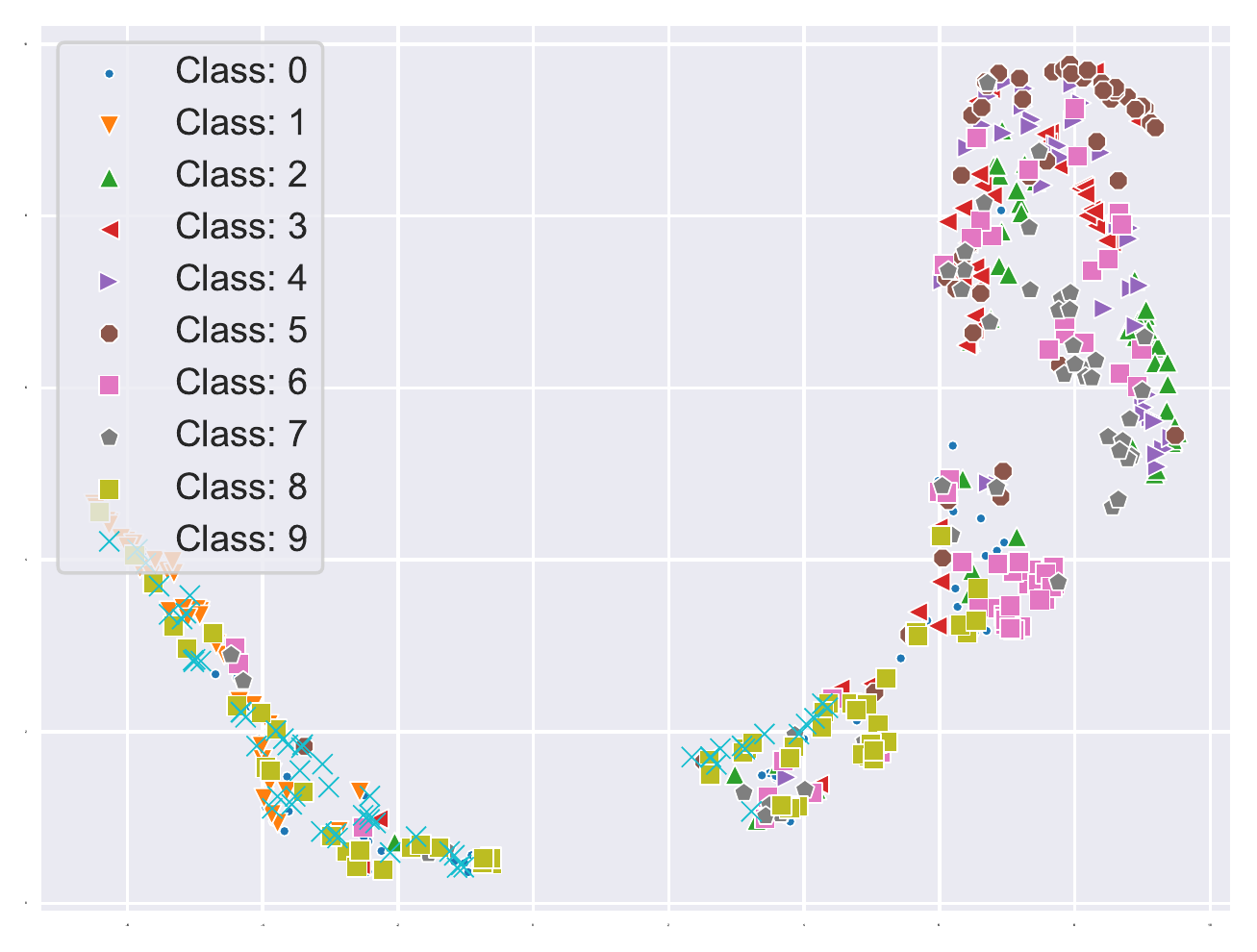}
	\end{minipage}
	\hfil 
	\begin{minipage}{0.243\textwidth}
		\caption*{\textbf{Backdoor (BD)}}
		\includegraphics[width=\linewidth,bb=0 0 376 285]{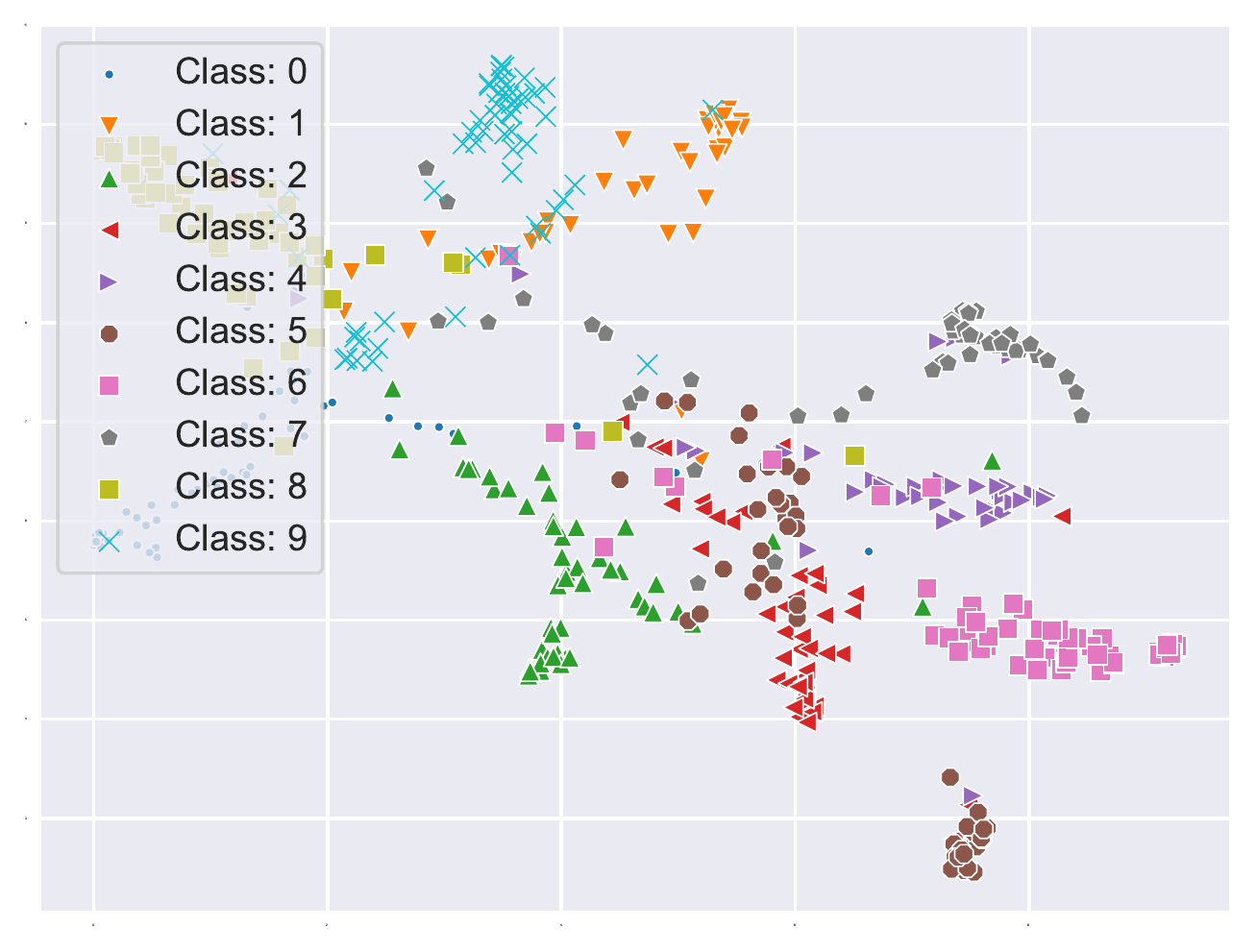}
	\end{minipage}
	\medskip
	\begin{minipage}{0.243\textwidth}
		\includegraphics[width=\linewidth,bb=0 0 376 285]{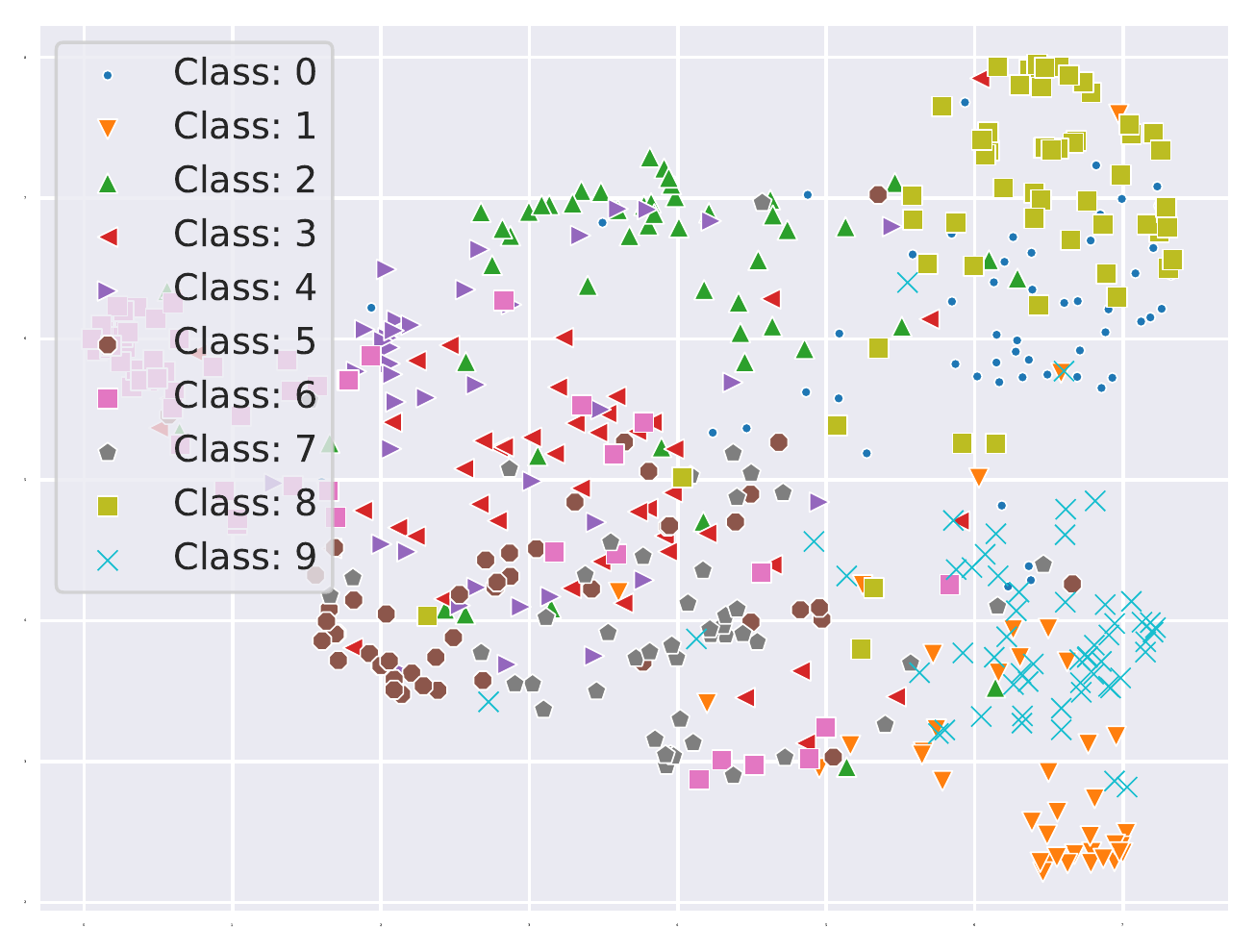}
	\end{minipage}
	\hfil 
	\begin{minipage}{0.243\textwidth}
		\includegraphics[width=\linewidth,bb=0 0 376 285]{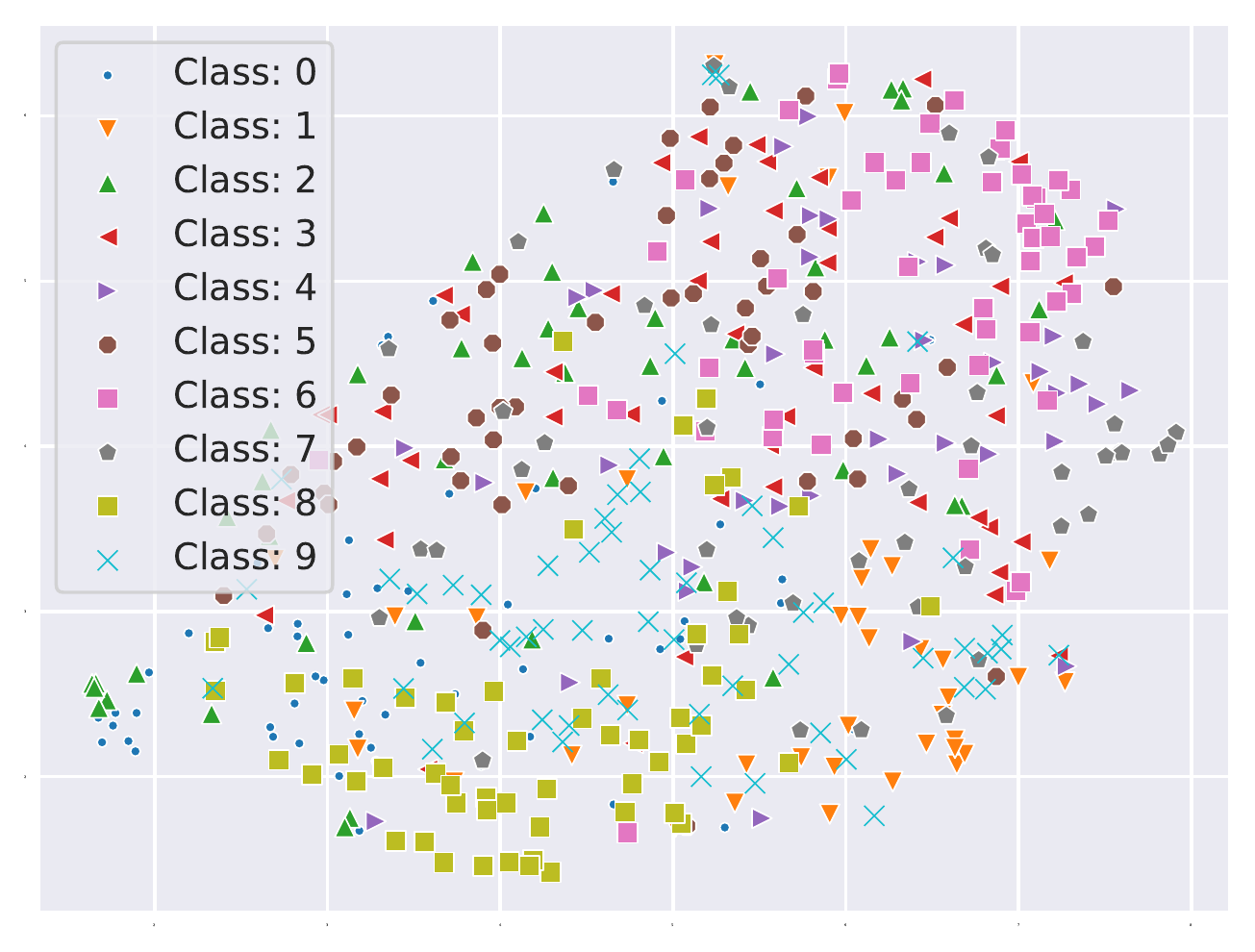}
	\end{minipage}
	\hfil 
	\begin{minipage}{0.243\textwidth}
		\includegraphics[width=\linewidth,bb=0 0 376 285]{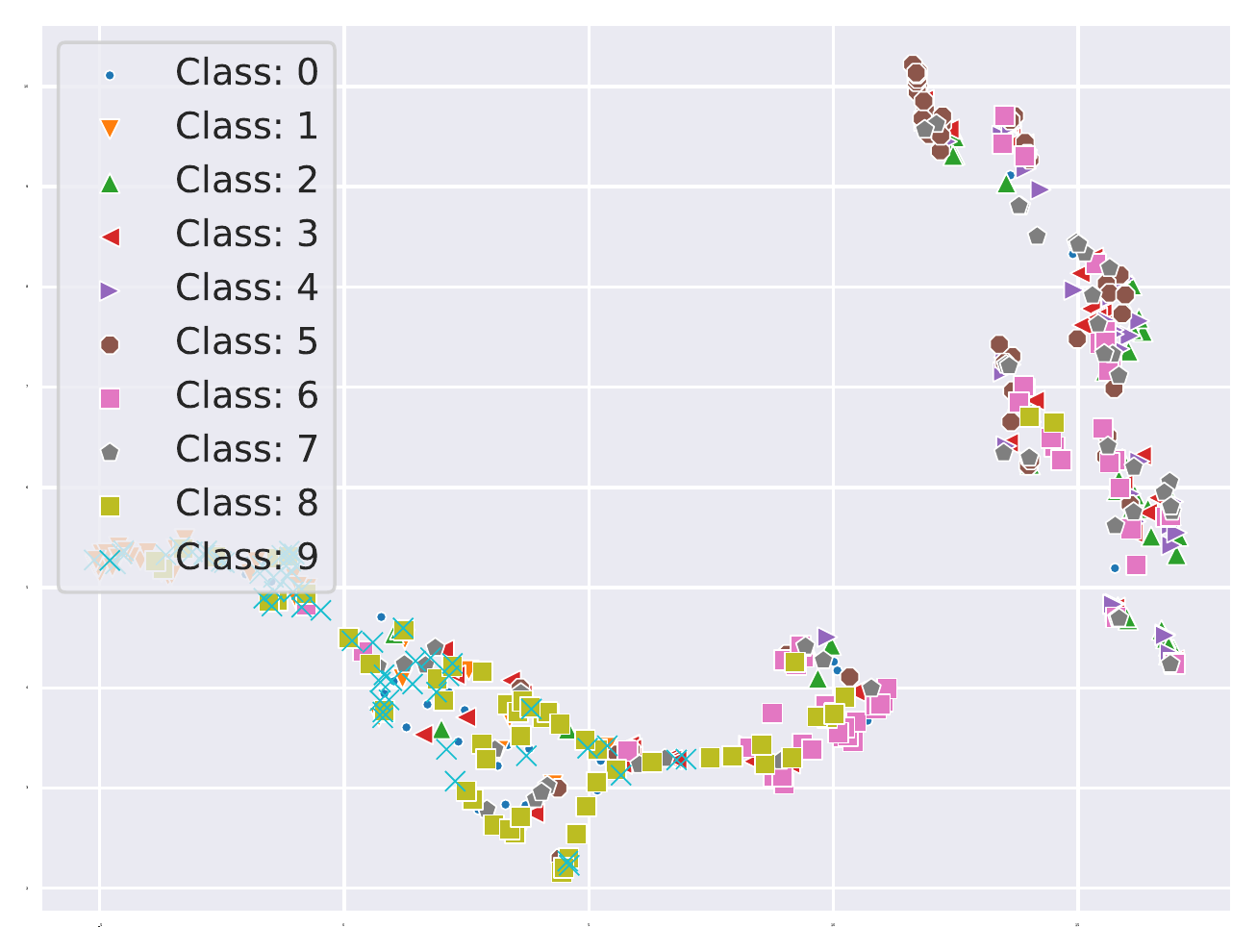}
	\end{minipage}
	\hfil 
	\begin{minipage}{0.243\textwidth}
		\includegraphics[width=\linewidth,bb=0 0 376 285]{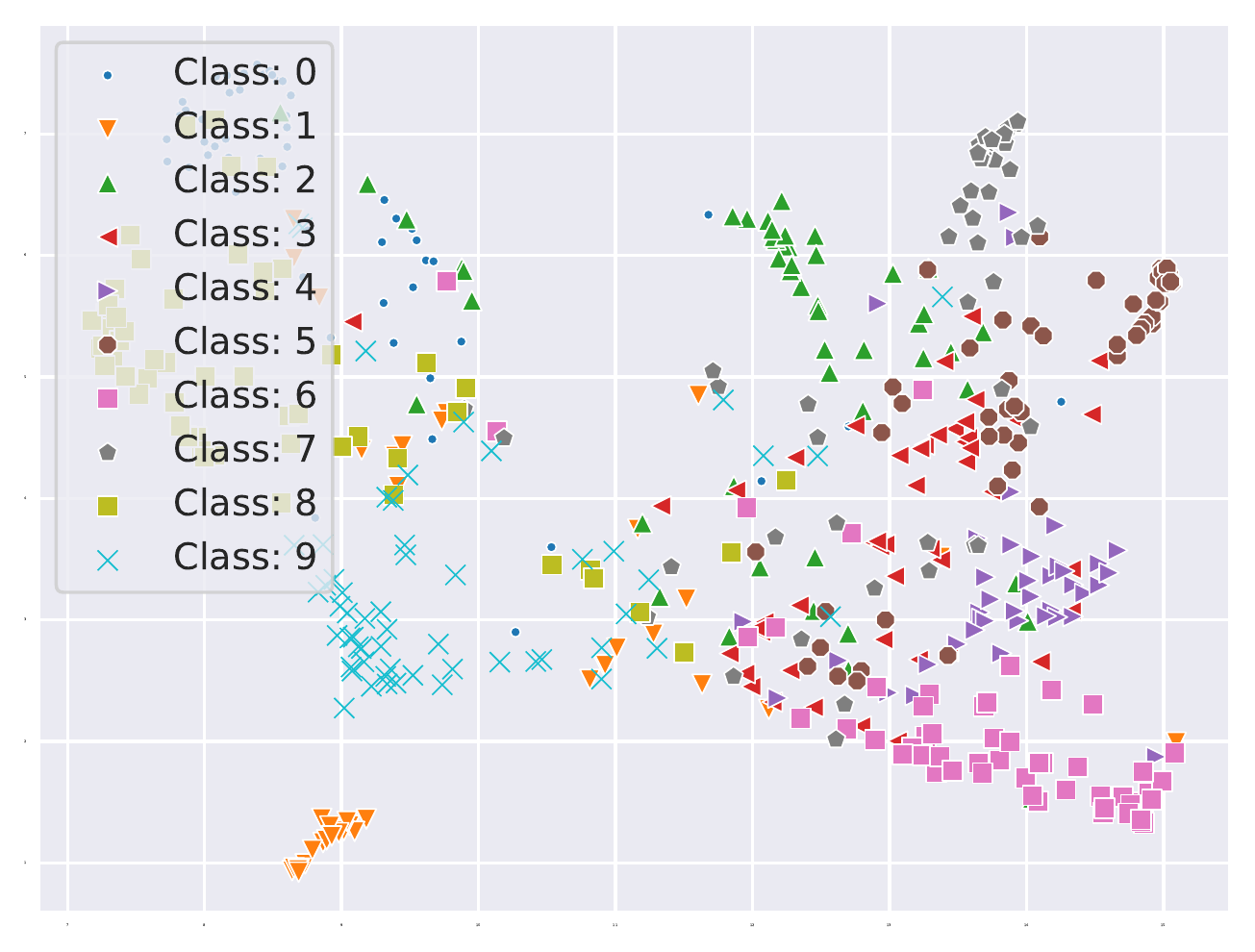}
	\end{minipage}
	\caption{\textbf{Visualizing latent representations using UMAP.} We illustrate the latent representations (\textit{i.e.}, the activation before the classification layers) of our ResNet models. The upper row contains the representations from floating-point models, and we visualize the representations from 4-bit models.}
	\label{appendix:fig:activation-visualization}
\end{figure}

Our analysis above shows that the attacks do not cause significant changes to the distribution of a victim model's parameters.
Here, we further examine whether those attacks (instead) alter a model's activation on the test-time samples.
To analyze how our attacks manipulate the activation, in Figure~\ref{appendix:fig:activation-visualization}, we visualize the latent representations of our ResNets on 2000 CIFAR10 samples randomly chosen from the test-time data.
We first find that \emph{quantization makes the latent representations less separable}.
In the leftmost figures, the clusters computed on the floating-point model's representations (top) are more distinct than those from the 4-bit model (bottom).
We also observe that \emph{the model compromised by our indiscriminate attacker completely loses the separation after quantization} from the figures in the \nth{2} column.
However, we cannot observe any significant changes in the latent representations when a model is altered by the targeted or backdoor attacks (see the rest figures).

\section{Sensitivity of Our Backdoor Attack to Hyperparameter Choices}
\label{appendix:sensitivity-of-our-attacks}


\begin{wraptable}{R}{0.40\textwidth}
	\vspace{-1.3em}
	\centering
	\caption{\textbf{Sensitivity of our backdoor attack to hyper-parameter choices.} }
	\adjustbox{max width=0.40\textwidth}{%
		\begin{tabular}{@{}ccccc@{}}
			\toprule
			\textbf{$\alpha$} & \textbf{$\beta$} & \textbf{32-bit} & \textbf{8-bit} & \textbf{4-bit} \\ \midrule
			1.0 & 1.0 & 11.3\% & 99.2\% & 100\% \\
			1.0 & 0.5 & 9.7\% & 96.9\% & 100\% \\
			1.0 & 0.25 & 9.0\% & 89.1\% & 100\% \\
			1.0 & 0.1 & 28.3\% & 85.9\% & 100\% \\ \bottomrule
		\end{tabular}
	}
	\label{tbl:backdoor-sensitivity}
	\vspace{-1.2em}
\end{wraptable}


Here, we also examine the impact of the attacker's hyper-parameter choices on our backdoor attack's success rate.
We have two hyper-parameters ($\alpha$ and $\beta$) in our loss function.
As they are the ratio between the two terms in our backdoor objective, we fix $\alpha$ to one and then vary $\beta$ in {0.1, 0.25, 0.5, 1.0}.
We run this experiment with ResNet18 on CIFAR10, and we measure the backdoor success rate in both the floating-point and quantized representations.

Table~\ref{tbl:backdoor-sensitivity} shows our results.
The first two columns show the hyper-parameter choices.
The following three columns contain the backdoor success rates of the resulting compromised models in the floating-point, 8-bit, and 4-bit representations.
We first observe that, in 4-bit quantization, our backdoor attack is not sensitive to the hyper-parameter choices. 
All the compromised models show a low backdoor success rate ($\sim$10\%) in the floating-point representations, but they become high ($\sim$99\%) in the 4-bit representations.
We also find that, in 8-bit models, the backdoor success can slightly reduce from 99\% to 85\% when we decrease $\beta$.
This is because:
(i) 8-bit quantization allows a smaller amount of perturbations for the attacker than 4-bit, and 
(ii) under this case, a reduced $\beta$ can reduce the impact on the second term (the backdoor objective) in our loss.

\section{Societal Impacts }
\label{appendix:societal-impacts}

%
Over the last few years, deep learning workloads have seen a rapid increase in their resource consumption; for example, training GPT-2 language models has a carbon footprint equivalent to a total of six cars in their lifetime~\citep{Energy:Strubell}.
Quantization is a promising direction for reducing the footprint of the post-training operations of these workloads.
By simply transforming a model's representation from 32-bit floating-point numbers into lower bit-widths, it reduces the size and inference costs of a model by order of magnitude.
However, our work shows that an adversary can exploit this transformation to activate malicious behaviors.
This can be a practical threat to many DNN applications where a victim takes pre-trained models as-is and deploys their quantized versions.
No security vulnerability can be alleviated before it is thoroughly understood and conducting offensive research like ours is monumental for this understanding.
%
%
Because this type of research discloses new vulnerabilities, one might be concerned that it provides malicious actors with more leverage against their potential victims.
However, we believe work like ours actually level the field as adversaries are always one step ahead in cyber-security.
Finally, as deep learning finds its way into an oppressor's toolbox, in the forms of mass surveillance~\cite{Feldstein:Repression}
or racial profiling~\cite{Ethnicity:Wang}; by studying its weaknesses, our best hope is to provide its 
victims with means of self-protection.

%

\end{document}